\documentclass{article}
\pdfoutput=1

\usepackage[nonatbib,preprint]{neurips_2024}

\usepackage[utf8]{inputenc} 
\usepackage[T1]{fontenc}    
\usepackage{url}            
\usepackage{booktabs}       
\usepackage{amsfonts}       
\usepackage{nicefrac}       
\usepackage{microtype}      
\usepackage{xcolor}         

\usepackage{times}
\usepackage{soul}
\usepackage{url}
\usepackage[small]{caption}
\usepackage{graphicx}
\usepackage{amsmath}
\usepackage{amsthm}
\usepackage{booktabs}
\usepackage{algorithm}
\usepackage{algorithmic}
\usepackage{multirow}
\usepackage[caption=false]{subfig}
\usepackage{amssymb}
\usepackage{booktabs}
\usepackage{color}
\usepackage{colortbl}
\usepackage{makecell}
\usepackage{array}
\usepackage{pifont}       
\usepackage{bbding}  
\usepackage{wrapfig}

\definecolor{mygray}{gray}{.9}
\usepackage[accsupp]{axessibility}
\usepackage[pagebackref=true,breaklinks=true,letterpaper=true,colorlinks,bookmarks=false]{hyperref}

\title{Rethinking Remote Sensing Change Detection With A Mask View}

\author{%
Xiaowen Ma$^{1,2}$, \quad Zhenkai Wu$^{1}$,  \quad Rongrong Lian$^{1}$, \quad Wei Zhang$^{1*}$, \quad Siyang Song$^{3}$\thanks{Corresponding author.} \vspace{.5em} \\
  \small{$^1$Zhejiang University \quad $^2$ Huawei Noah’s Ark Lab \quad $^3$University of Leicester} \vspace{.5em} \\
\small\textbf{\url{https://github.com/xwmaxwma/rschange}}
}

\begin{document}

\maketitle
\begin{abstract}
  Remote sensing change detection aims to compare two or more images recorded for the same area but taken at different time stamps to quantitatively and qualitatively assess changes in geographical entities and environmental factors. 
  Mainstream models usually built on pixel-by-pixel change detection paradigms, which cannot tolerate the diversity of changes due to complex scenes and variation in imaging conditions. To address this shortcoming, this paper rethinks the change detection with the mask view, and further proposes the corresponding: 1) meta-architecture CDMask and 2) instance network CDMaskFormer. Components of CDMask include Siamese backbone, change extractor, pixel decoder, transformer decoder and normalized detector, which ensures the proper functioning of the mask detection paradigm. Since the change query can be adaptively updated based on the bi-temporal feature content, the proposed CDMask can adapt to different latent data distributions, thus accurately identifying regions of interest changes in complex scenarios. Consequently, we further propose the instance network CDMaskFormer customized for the change detection task, which includes: (i) a Spatial-temporal convolutional attention-based instantiated change extractor to capture spatio-temporal context simultaneously with lightweight operations; and (ii) a scene-guided axial attention-instantiated transformer decoder to extract more spatial details. State-of-the-art performance of CDMaskFormer is achieved on five benchmark datasets with a satisfactory efficiency-accuracy trade-off.

\end{abstract}

\section{Introduction}
\label{sec:intro}

The rapid development of remote sensing platforms and sensors has opened up the possibility of detailed observation of Earth's surface features. Remote sensing change detection (RSCD) compares and analyzes remote sensing images recorded from the same area but with different time stamps to determine changes in the surface. This technology has a wide range of applications in the fields of environmental monitoring \cite{deforestation}, resource management \cite{landuse2} and urban planning \cite{urban}. 

Early RSCD methods were frequently built on hand-crafted algebra \cite{cha}, transformation \cite{cva}, or classification \cite{rand} strategies to detect changed pixels, which are often ineffective on high-resolution remote sensing images \cite{cva}. As a result, Convolutional Neural Networks (CNNs)-based RSCD methods \cite{fc-siam} have been widely developed due to their task-specific feature extraction capabilities. They focus on increasing the receptive field of the model by stacking more convolutional layers \cite{snunet}, dilate convolution \cite{zhang2018triplet}, or attention mechanisms \cite{lgpnet}, which improve the representing capability of the extracted features for describing target changes. Recently, transformer-based models \cite{bit,changeformer,sarasnet} are popular for the RSCD, as their powerful global context modeling capability is beneficial to capture the relational dependencies of bi-temporal features in spatial or temporal scales. However, all of these CNN/Transformer-based solutions are built on the pixel-by-pixel change detection paradigm (CDPixel) that depends on fixed semantic prototypes to detect changes of interest. Therefore, it cannot tolerate different data distributions of changes produced by complex scenes and variation in imaging conditions (e.g., weather, light, season, and frequent irrelevant changes caused by human actions), as shown in Fig. \ref{fig:intro}. Detailed analysis are provided in the Appendix. 

\begin{figure}[t]
	\centering
	\includegraphics[width=0.99\textwidth]{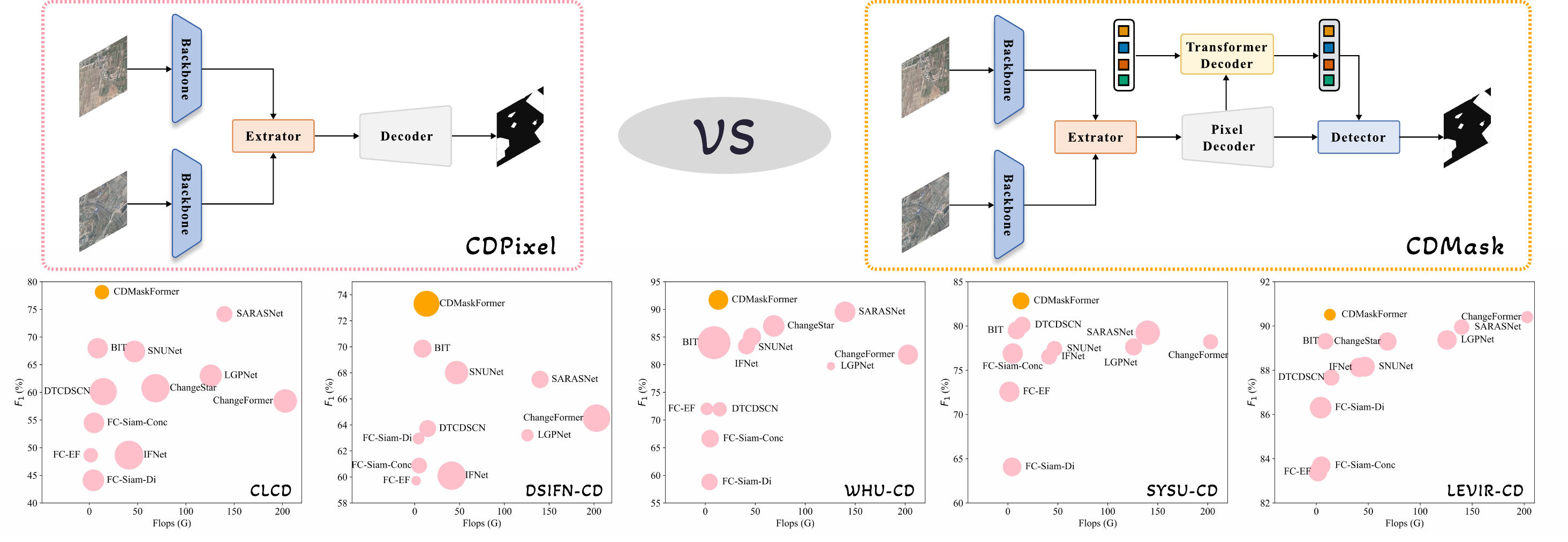}
	\caption{Visualization of (a) architecture Comparison of CDPixel and CDMask and (b) Performance-Computing curves on five benchmark datasets. FLOPs are calculated using an input size of $256 \times 256$. The experiments are carried out five times with different random seeds, with the center of the circle indicating the median value of the performance and the size of the radius of the circle indicating the standard deviation of the performance. It can be observed that CDMaskFormer achieves state-of-the-art results and the most satisfactory trade-off between change detection performance and computational complexity on five benchmark datasets.}
	\label{fig:res}
\end{figure}

Inspired by recent successes of mask classification models (or DETRs) for object detection \cite{detr,groupdetr,rankdetr,rtdetr} and image segmentation \cite{maskformer,mask2former,pem}, we revisit the change detection task from a mask view. Specifically, we propose a novel meta-architecture CDMask which introduces learnable change queries to predict a set of binary masks based on the feature content of bi-temporal images, and then classify the masks to determine whether a change of interest has occurred. It is made up of a Siamese backbone, a change extractor, a pixel decoder, a transformer decoder and a normalized detector (Fig.~\ref{fig:intro} (a)) that ensures the proper functioning of the mask detection paradigm (analysis is provided in Sec. \ref{cdmask-sec}). Importantly, our CDMask possesses two distinct advantages: (i) it can better tolerate the diversity of changes by adaptively generating a change mask based on the features of the bi-temporal image; and (ii) it is compatible with a variety of advanced DETRs frameworks \cite{mask2former,rtdetr,rankdetr} by requiring only minor modifications, which is conducive to the establishment of the a unified-style framework for various remote sensing image analysis tasks.

However, the existing components of CDMask have obvious shortcomings for change detection, such as the change extractor cannot effectively model the spatio-temporal global context, and the transformer decoder has insufficient fine-grained recognition capability. Therefore, we further propose a novel instance network CDMaskFormer customized from CDMask, aiming to achieve superior performance while maintaining lightweight. It is built on two main designs: (i) a novel spatial-temporal convolutional attention mechanism that instantiates the change extractor, which requires minimal computational cost to model the global spatio-temporal context to extract higher quality change representations; and (ii) a scene-guided axial attention mechanism to instantiate the transformer decoder to mine more detailed information from higher resolution change representations. 
 
The main contributions of this paper can be summarized as follows.
\begin{itemize}

\item We analyze the shortcomings of existing pixel-by-pixel change detection paradigms and propose to generate adaptive change masks via learnable change queries. To the best of our knowledge, it is the first change detection method based on mask classification and provides a new paradigm for the design of subsequent work.

\item We propose the meta-architecture CDMask, which requires only minor changes to be compatible with various state-of-the-art DETRs frameworks. In particular, we design a Normalized detector in CDMask, which is a key component for CDMask to work well.

\item We propose the instance network CDMaskFormer, whose components are highly customizable, where a novel Spatial-temporal convolutional attention and scene-guided axial attention mechanism are included to instantiate the change extractor and transformer decoder. Our novel CDMaskFormer outperformed previous state-of-the-art models on five RSCD benchmark datasets.

\end{itemize}

\begin{figure}[t]
	\centering
	\includegraphics[width=0.9\textwidth]{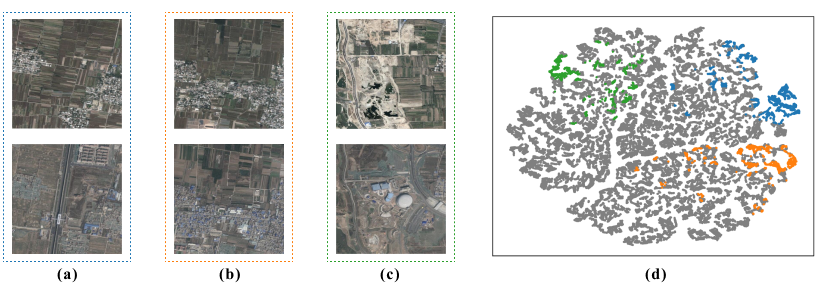}
	\caption{Visualization of the latent feature distribution for changes of interest in different bi-temporal images of BIT \cite{bit}, which is an instance of CDPixel. Blue, orange and green colors represent features belonging to the changes of interest in the image pairs (a), (b) and (c), and gray color indicates unchanged features, respectively.}
	\label{fig:intro}
\end{figure}

\section{Related Works}
\textbf{Remote Sensing Change Detection:} Existing remote sensing image change detection tasks can be categorized into two types: convolution-based and transformer-based solutions. Convolution-based methods can be broadly categorized into two types based on the fusion stage of bi-temporal information. Image-level methods \cite{imagecd1,imagecd3} connect the bi-temporal image as a single input to a semantic segmentation network. The feature-level methods \cite{lgpnet,changestar,dmatnet} combine bi-temporal features extracted from the neural network and makes change decisions based on the fused features. Recent studies frequently aim to improve the feature differentiation capability of neural networks by designing multilevel feature fusion structures for better contextual modeling in terms of spatial and temporal scales \cite{bit,dmatnet}. Furthermore, in high-resolution remote sensing imagery, context modeling is key to identifying changes in interest due to the complexity of objects in the scene and changes in image conditions. To increase the size of the receptive field, existing approaches include employing deeper CNN models \cite{deep1,deep2,deep3}, using dilation convolution \cite{dilate1}, and applying attention mechanisms \cite{att1,att2,att3}. Transformer-based solutions usually show superior performances due to their ability to utilize global relationships between pixels in space-time. ChangeFormer \cite{changeformer} applies the Transformer directly to accomplish the change detection task. However, the large computational complexity limits its application. BIT \cite{bit} proposes to semantically tokenize the features from the convolutional network and use the Transformer encoder to model the context in a compact label-based spacetime, and then, the learned context-rich token is fed back into the pixel space, where it is passed through the Transformer decoder to refine the original features. SARASNet \cite{sarasnet} emphasizes the interaction information of two different images and proposes technical tools such as relation-awareness, scale-awareness, and cross-transformation to solve the scene-switching detection problem more effectively.

\textbf{Mask-level Classification:} DETR \cite{detr} is the first Transformer-based end-to-end target detection network that reasons about the relationship between objects and the global image context given a set of object queries and outputs the final prediction set in parallel. Inspired by DETR, MaskFormer \cite{maskformer} proposes the first mask classification model for semantic segmentation, which breaks the paradigm of pixel-by-pixel classification by learning the object queries to classify the masks. Mask2Former \cite{mask2former} proposes to constrain the cross-attention within the prediction mask region to accelerate the model convergence. FASeg \cite{faseg} and MP-Former \cite{mpformer} propose dynamic positional query and truth mask bootstrapping to further improve model performance respectively. 
Inspired by the mask classification models in the fields of object detection \cite{focusdetr,dacdetr,rankdetr,rtdetr,dino} and image segmentation \cite{maskformer,maskdino,pem}, we try to rethink change detection from the perspective of masks. Therefore, we design the meta-architecture CDMask, which can be abstracted as a Siamese backbone, a change extractor, a pixel decoder, a transformer decoder, and a normalized detector, which is specifically proposed to ensure the proper functioning of the mask detection paradigm. Compared to general DETRs \cite{detr,maskformer}, CDMask makes many architectural changes to accommodate the change detection task. However, it is important to note that many of the advanced designs for these tasks such as masked attention \cite{mask2former} and deformable attention \cite{deformabledetr} can be used to instantiate transformer decoders and pixel decoder. Thus, CDMask has good compatibility with many good DETRs frameworks.

\section{Methodology}
Existing mainstream models usually adopt a pixel-by-pixel change detection paradigm, which cannot tolerate diverse changes due to scene complexity and changing imaging conditions. Inspired by DETRs \cite{detr,mask2former}, we revisit the change detection task from the mask perspective and further propose: 1) a meta-architecture CDMask; and 2) a instance network (called CDMaskFormer) customized from the meta-architecture CDMask, aiming at achieving superior change detection performance with lighter operations.

\subsection{CDMask}
\label{cdmask-sec}
Fig. \ref{fig:res} illustrates the main structure of our CDMask, which contains a Siamese backbone, a change extractor, a pixel decoder, a transformer decoder and a normalized detector. Given input images $\mathcal{T}_1$ and $\mathcal{T}_2$, the bi-temporal features $\mathcal{F}$ are first extracted through a pair of weight-shared backbones. Then, the change extractor is applied to fuse the bi-temporal features to produce the high quality change representations $\mathcal{R}$. The change queries $\mathcal{R}_q$ are then adaptively updated depending on different data distributions corresponding to image content. Finally, the updated change queries $\mathcal{R}_q$ and the change representations $\mathcal{R}$ are input to the normalized detector and the output mask is obtained by up-sampling.

\textbf{Siamese backbone.} Following the previous pixel-by-pixel change detection works \cite{bit,snunet}, our CDMask extracts bi-temporal features based on weight-sharing Siamese backbones, where a hierarchical backbone is employed to facilitate better modeling of multi-scale geospatial objects in remote sensing images. This is inpired by \cite{resnet,swin,swinv2} that hierarchical backbones are preferred in most cases over non-hierarchical backbones such as plain Vision Transformers \cite{vit,beit}. 

\textbf{Change extractor.} The change extractor aims to fuse diachronic features and generate change representations for updating change queries. Note that the change extractor in CDPixel (i.e., the module in CDPixel that fuses the diachronic features, such as splicing \cite{fc-siam}, element-level subtraction \cite{fc-siam}, dense connectivity \cite{snunet}, cross-attention \cite{dminet} and state-space modeling \cite{changemamba}) can be directly applied to instantiate the corresponding components of the CDMask.

\textbf{Pixel decoder.} The pixel decoder is introduced to extract multiscale features, allowing for accurate detection of change areas. The input to the pixel decoder is the multi-scale change representations due to preprocessing by the change extractor. Therefore, common pixel decoder designs such as deformable attention \cite{mask2former,deformabledetr} and feature pyramids \cite{pem} can be directly compatible in CDMask.

\textbf{Transformer decoder.} Referring to the design of DETRs, our CDMask introduces a learnable change query to interact with the change representation for information, and thus adaptively generates a change mask. Similarly, the existing transformer decoder design of DETRs is directly compatible with CDMask. The difference is that the number of change queries in CDMask is smaller (often less than 10, defaulting to 5 in this paper), whereas the number of queries introduced by DETRs is often no less than (or even much larger than) 100.This can be explained by the different requirements of the tasks. Previous DETRs often required the detection or segmentation of a large number of instance targets, thus requiring a large number of queries to generate enough candidate masks. In change detection, only regions of changing interest need to be acquired, which can be abstracted as a binary classification problem. Therefore, fewer change queries can generate accurate change masks.

\begin{figure}[t]
	\centering
        \includegraphics[width=0.9\textwidth]{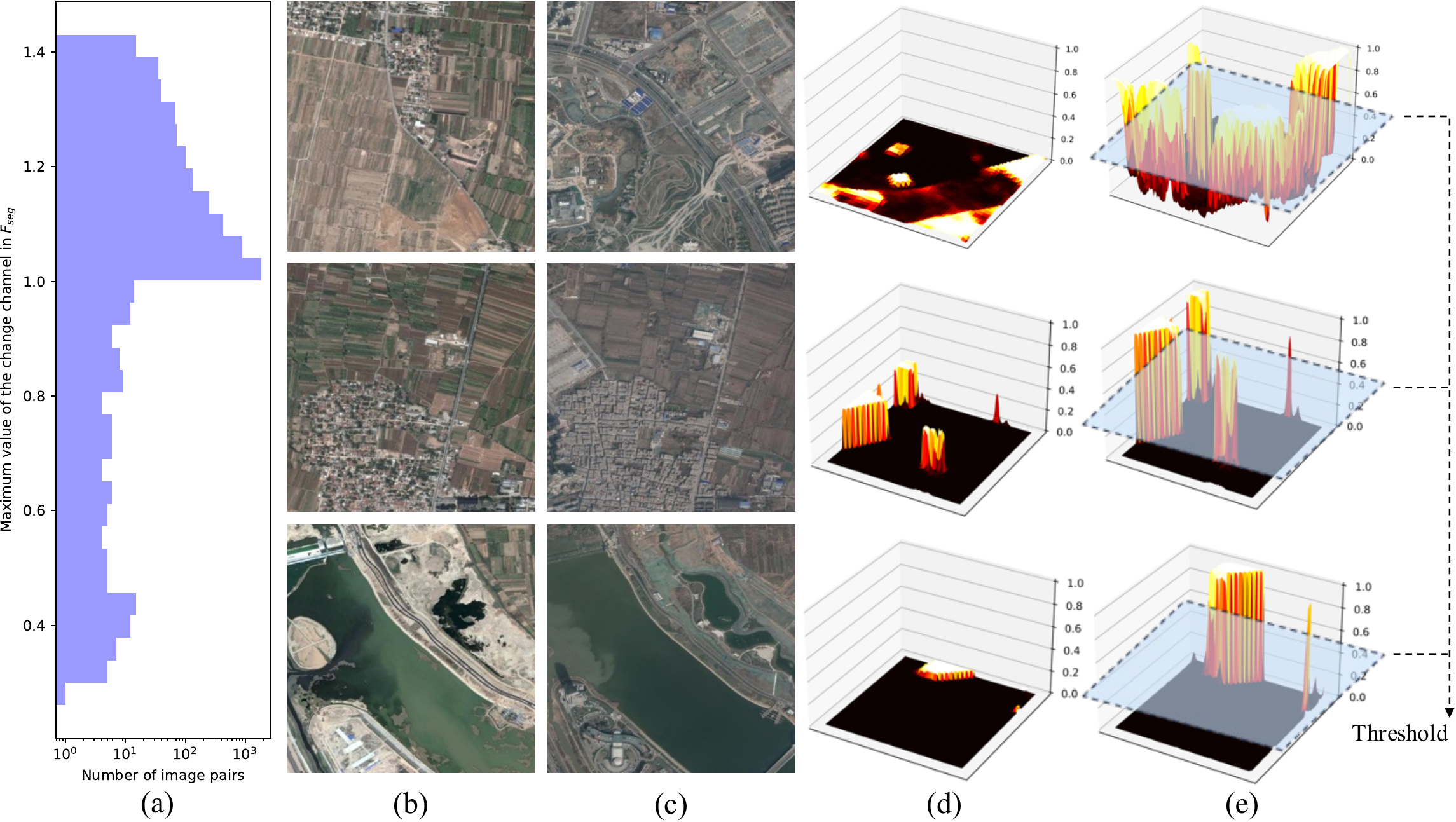}
	\caption{Description of the normalized detector. The range of output values is different for different input images. (a) is the statistics of the maximum value of the change channel on the DSIFN-CD dataset. (b) and (c) are example bi-temporal images. (d) and (e) are heat maps of the values before and after Normalized, respectively. We introduce min-max Normalized to map the data to between 0 and 1, so that detect changes based on a fixed threshold.}
	\label{fig:detector}
\end{figure}

\textbf{Normalized detector.}
The original mask classification model \cite{maskformer,mask2former} follows three steps for pixel classification: First, given the change prototype $\mathcal{R}_p \in \mathbb{R}^{m\times C}$ output from the detail-enhanced decoder, a multilayer perceptron (MLP) is applied to generate the category prediction $\mathcal{O}_{cls} \in \mathbb{R}^{m\times K}$, where $m$ and $C$ denotes the number and channels of the prototypes, respectively; $K$ denotes the number of categories, which is $2$ in the RSCD task, i.e., change class and unchange class ($\varnothing$). 
Second, another MLP is applied to process $\mathcal{R}_p$ and then multiply the result obtained with $\mathcal{R}_4$ to produce the mask predictions $\mathcal{O}_{mask} \in \mathbb{R}^{m\times H \times W}$, where $H$ and $W$ denote the height and width of the masks. 
Third, $\mathcal{O}_{mask}$ is multiplied with $\mathcal{O}_{cls}$ and then drop the value of $\varnothing$,
\begin{equation}
\label{drop}
    \mathcal{F}_\text{seg} = ( \mathcal{O}_{cls} \otimes \mathcal{O}_{mask} ) _{\text{drop}\ \varnothing}.
\end{equation}

However, the above classification scheme suffer from two main issues when applying it to the RSCD task:
1) if the value of $\varnothing$ decreased, $\mathcal{F}_\text{seg} \in \mathbb{R}^{H \times W \times 1}$ has only one output channel and cannot determine which pixels belong to the changes of interest; and
2) if the value of $\varnothing$ is retained, most change queries tend to represent $\varnothing$, which leads to the value of unchange class that tend to be larger than the value of the change class for pixels in $\mathcal{F}_\text{seg}\in \mathbb{R}^{H \times W \times 2}$. Thus, the detector can not output the required change mask.

Consequently, it is necessary to re-design the detector that can correctly determine the category of each target pixel, thus adapting it to the RSCD task. We analyse the value of $\mathcal{F}_\text{seg}$ and obtain the following \textbf{two key observations} (Detailed discussed in Appendix): 
1) The response values of changed pixels are significantly higher than unchanged pixels on the channels of the change class, which makes them easy to be detected; and 
2) Different images possess different ranges on the channels of the change class, as shown in Figure \ref{fig:detector}. For example, for the value of the channel corresponding to change class in $\mathcal{F}_\text{seg}$, some images are greater than 1, some are less than 0.5 or even lower. Inspired by these observations, we propose a novel normalized detector. As shown in Eq. \ref{drop}, it retains the operation of dropping $\varnothing$ to obtain $\mathcal{F}_\text{seg}$. Then, it maps the output values of the change channels to between 0 and 1 based on min-max normalization as:
\begin{equation}
    \mathcal{F'}_\text{seg} = \text{Normalize}_{\text{(min-max)}}(\mathcal{F}_\text{seg}),
\end{equation}
This allows us to detect changes based on a fixed threshold $\mathcal{T}$, i.e.,
\begin{equation}
    \hat{Y}=\left\{ \begin{array}{l}
    	1,\ \mathcal{F}_\text{seg}'>\mathcal{T}\\
    	0,\ \text{else}\\
    \end{array} \right. 
\end{equation}

\textbf{Loss function.} CDMask uses binary cross-entropy loss and dice loss~\cite{dice} to supervise the mask. In addition, cross-entropy loss is applied to supervise the class probabilities. The final loss is a combination of mask loss and classification loss: $\mathcal{L} = \lambda_{ce}\mathcal{L}_{ce} + \lambda_{dice}\mathcal{L}_{dice} + \lambda_{cls}\mathcal{L}_{cls}$.

\begin{figure*}[t]
	\centering
	\includegraphics[width=0.9\textwidth]{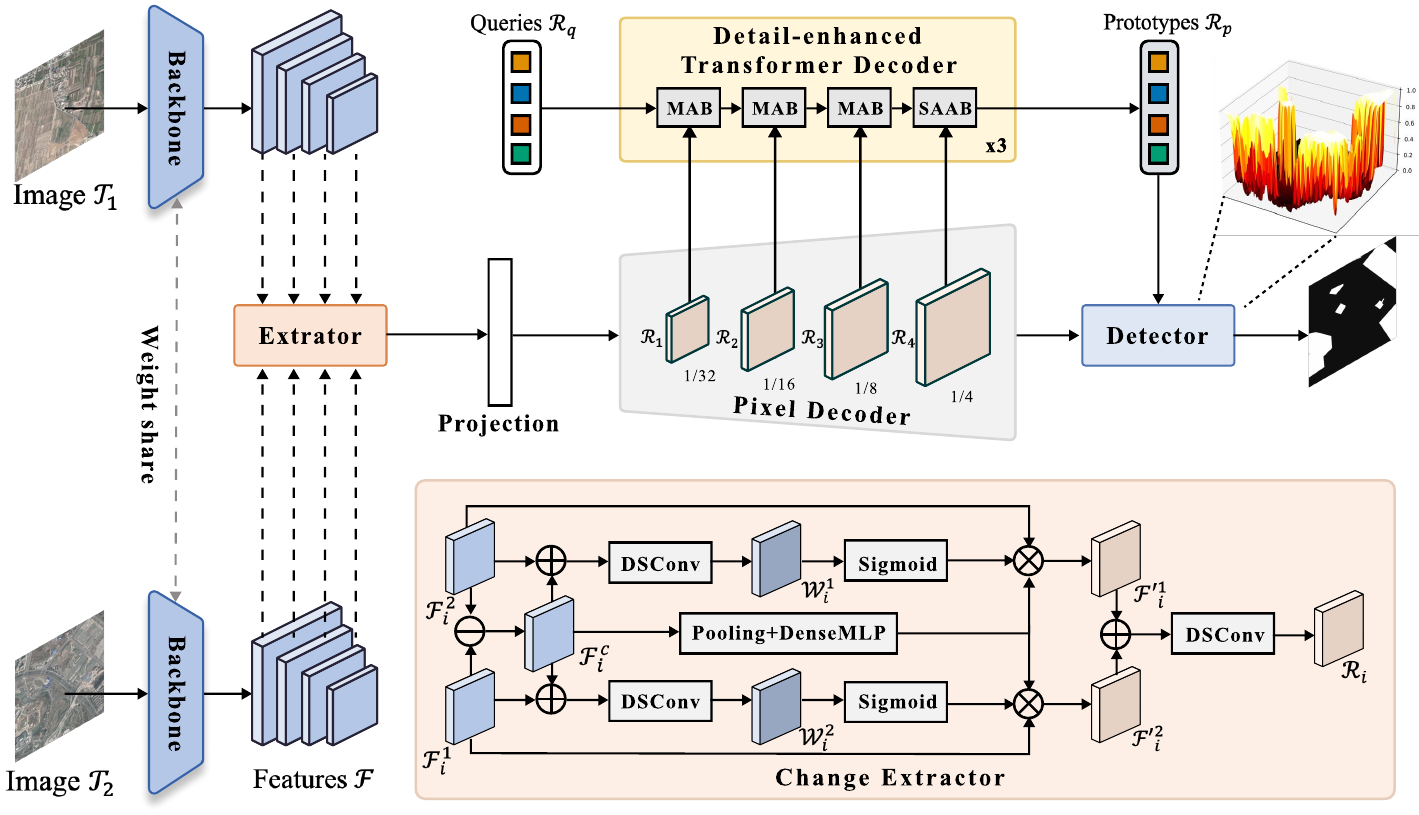}
	\caption{Architecture of the CDMaskFormer. Given the bi-temporal images $\mathcal{T}_1$ and $\mathcal{T}_2$, a pair of weight-shared backbones is applied to obtain the features $\mathcal{F}$. For each layer of bi-temporal features $\mathcal{F}_i^1$ and $\mathcal{F}_i^2$, a change extractor is passed through to obtain the change representations $\mathcal{R}_i$, which is passed through a projection matrix to unify channels, and then refined by a pixel decoder. Then, the randomly initialized change queries $\mathcal{R}_q$ and the refined change representations are input to the detail-enhanced decoder for information interaction. The change queries are updated by the bi-temporal feature content. Finally, the obtained change prototypes $\mathcal{R}_p$ with $\mathcal{R}_4$ are input into the normalized detector to obtain the output mask. Note that DWConv and PWConv denote depth-wise convolution and point-wise convolution, respectively, and DMLP denotes dense multilayer perceptron, which will be described in Section 3.2.}
	\label{fig:whole}
\end{figure*}

\subsection{CDMaskFormer}
To further exploit the potential of CDMask, we propose the instantiation network CDMaskFormer, whose structure is shown in Fig.~\ref{fig:whole}. The components of CDMaskFormer are highly customized for the change detection task, aiming to try to achieve better performance with more lightweight operations. We use deformable attention \cite{deformabledetr} to instantiate the pixel decoder and apply a lightweight backbone \cite{seaformer}. The highlights of CDMaskFormer mainly lie in two designs, 1) the use of Spatial-temporal convolutional attention to instantiate the change extractor and 2) the use of scene-guided axial attention to instantiate the transformer decoder, which is described in next subsections.

\subsubsection{Change Extractor Instantiated with Spatial-temporal Convolutional Attention}

This subsection proposes a lightweight change extractor to fuse bi-temporal features at different scales to obtain high-quality change representations. It can facilitates the subsequent information interaction between change queries and change representations. Specifically, our change extractor is built on cheap convolution attention simultaneously capturing the spatio-temporal range of contexts to selectively enhance changes of interest and suppress interference from irrelevant changes. 

Given the bi-temporal features $\mathcal{F}_i^1$ and $\mathcal{F}_i^2$ extracted from different spatial scales, we first apply a element-wise subtraction to generate the coarse change representations $\mathcal{F}_i^c$. Then, we concatenate $\mathcal{F}_i^c$ with the bi-temporal features separately. These features are fused after the deep separable convolution \cite{dsconv} to generate weight matrixs $\mathcal{W}_i^1$ and $\mathcal{W}_i^2$ as:
\begin{equation}
    \mathcal{W}_i^t = \text{DSConv}(\mathcal{F}_i^c \oplus \mathcal{F}_i^t), t\in \{1,2 \}.
\end{equation}
Note that we choose deep separable convolution (DSConv) for two reasons: 1) DSConv can greatly reduce the number of parameters of the model and decrease the computational consumption during the fusion process; and 2) depth-wise convolution of the DSConv has been proven to be effective in detail enhancement \cite{ssa}, which allows the model learning more details in the spatial dimension to generate more accurate weights. 

After that, we perform the sigmoid activation on $\mathcal{W}_i^1$ and $\mathcal{W}_i^2$ in order to enhance the features of change regions from the spatial dimension, i.e., $\mathcal{W'}_i^t = \sigma(\mathcal{W}_i^t), t\in \{1,2 \}$, 
where $\sigma$ denotes the sigmoid function. In addition, inspired by~\cite{senet,senetv2}, we model $\mathcal{F}_i^c$ in the temporal dimension to enhance features describing the changing regions in the bi-temporal images. Specifically, we first perform global average pooling of $\mathcal{F}_i^c$ and obtain the response values of the channels through a dense multilayer perceptron (DMLP) as: 
\begin{equation}
    \mathcal{W}_{i,j}^c = \phi_j(pooling(\mathcal{F}_i^c)),
\end{equation}
where $\phi_j$ denotes a DMLP, which consists of a linear mapping and a ReLU activation function. We then splice the outputs of DMLP along the channel dimension and go through a linear mapping and sigmoid function in order to obtain the weights of the channels,
\begin{equation}
    \mathcal{W'}_{i}^c=\sigma (\sum_j (\mathcal{W}_{i,j}^c)),
\end{equation}
where $\sum_j$ denotes concatenation along the channel. We employ DMLP to enhance the network's ability to capture channel patterns and global knowledge to enhance more discriminative feature of changing regions in the time dimension. 

Finally, we augment the bi-temporal features based on the weights obtained in the spatial and temporal dimensions, respectively, which are further processed by a DSConv to obtain the final change representations $\mathcal{R}_i$:
\begin{equation}
\label{fuse}
    \mathcal{R}_{i} = \text{DSConv}((\mathcal{W'}_{i}^c \otimes \mathcal{W'}_{i}^1 \otimes \mathcal{F}_i^1)\oplus(\mathcal{W'}_{i}^c \otimes \mathcal{W'}_{i}^2 \otimes \mathcal{F}_i^2)),
\end{equation}
As shown in Eq. \ref{fuse}, we modulate the fusion process of bi-temporal features based on cheap convolutional attention. Due to the simultaneous capture of the spatio-temporal context, the change extractor is able to significantly suppress the interference of irrelevant changes. We will discuss the effectiveness of the attention in spatial and temporal dimensions in Section 4.4.

\begin{figure*}[t]
	\centering
	\includegraphics[width=0.88\textwidth]{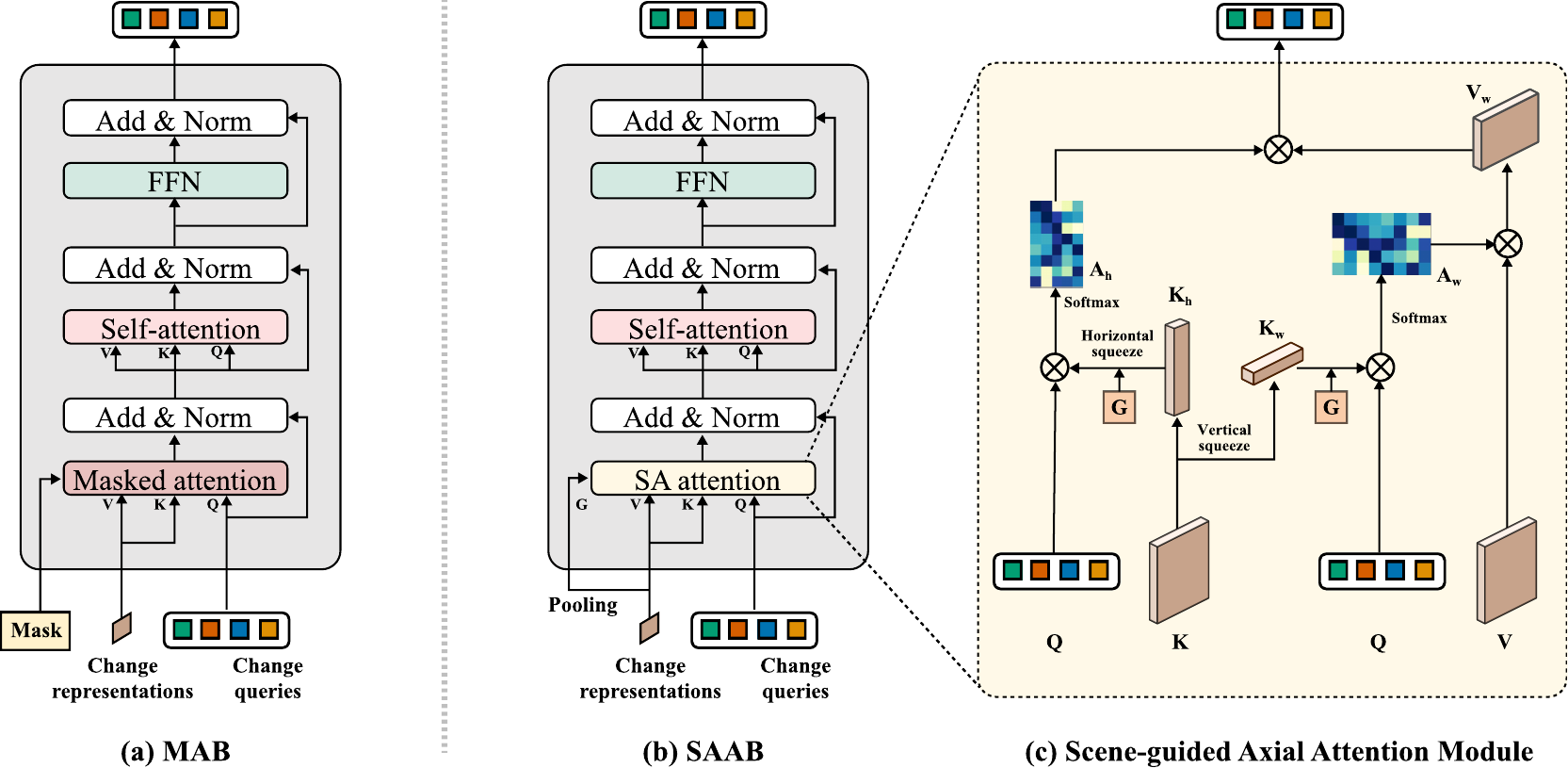}
	\caption{Structure of masked attention block (MAB), scene-guided axial attention block (SAAB) and scene-guided axial attention module. The SAAB use scene-guided axial attention to facilitate the informative interaction of change queries with high-resolution feature maps to mine further details.}
	\label{fig:saab}
\end{figure*}

\subsubsection{Transformer Decoder Instantiated with Scene-guided Axial Attention} 

Most regions are stay unchanged in a pair of bi-temporal images in RSCD tasks. As a result, more spatial details are needed to be captured in order to accurately detect the changed region. Therefore, we propose to instantiate the transformer decoder using a scene-guided axial attention mechanism to mine more detailed information from higher resolution representations of changes.
Specifically, we follow the design of Mask2Former \cite{mask2former}, i.e., for the 1/32, 1/16 and 1/8 resolutions, we employ a set of Masked Attention Blocks (MABs) \cite{mask2former} to guide model training and accelerate convergence. Fig.~\ref{fig:saab}(a) illustrates the structure of the MAB, which consists of a masked attention module, a self-attention module and a feed-forward network (FFN). Then, we follow the structure of the MAB to design Scene-guided Axial Attention Block (SAAB) whose core component is the scene-guided axial attention module. Here, we introduce the axial attention for two reasons: 1). there are many strip objects in remote sensing images, such as roads, rivers, and buildings. Thus, axial attention can be used as a  complement to ordinary cross-attention, which is conducive to extract strip-like features; 2). conducting cross-attention between object queries and high-resolution image features requires a large amount of memory consumption and computational cost, while our axial attention can reduce such computational consumption \cite{rcda}. 

However, the receptive field of original axial attention is restricted, which is not conducive to efficient cross-attention modeling. To deal with this issue, we introduce scene context to guide the interaction process between the query and the change representations. As shown in Fig.~\ref{fig:saab}(c), given the change query and the 1/4-resolution change representation $\mathcal{R}_4$, we first project them via a linear mapping to obtain $\mathcal{Q}$, $\mathcal{K}$ and $\mathcal{V}$. Then,  we compress $\mathcal{K}$ horizontally and vertically to obtain $\mathcal{K}_h$ and $\mathcal{K}_w$, respectively. In addition, we globally average-pool $\mathcal{R}_4$, following a linear mapping and sigmoid activation to obtain the scene context $G$. 
\begin{equation}
    G = \sigma(\text{Linear}(\text{Pooling}(\mathcal{R}_4))).
\end{equation}
Based on $G$, we weight $\mathcal{K}_h$ and $\mathcal{K}_w$ to regulate the inner product between queries and keys, and hence the affinity matrices $\mathcal{A}_h$ and $\mathcal{A}_w$, 
\begin{equation}
    \mathcal{A}_h = \text{Softmax}((G \otimes \mathcal{K}_h)\otimes \mathcal{Q}  ),\quad
    \mathcal{A}_w = \text{Softmax}((G \otimes \mathcal{K}_w)\otimes \mathcal{Q} ).
\end{equation}

In the practical implementation, we use a tandem axial attention scheme,
\begin{equation}
    \mathcal{Q}_{out} = \mathcal{A}_h \otimes \mathcal{V}_w = \mathcal{A}_h \otimes \mathcal{A}_w \otimes \mathcal{V},
\end{equation}
Finally, SAAB can facilitate change queries to extract more details from high-resolution change representations at the cost of adding a tiny amount of computation. In addition, the global context guided axial attention scheme significantly improves the model's ability to extract strip-like features.

\setlength{\tabcolsep}{3pt}
\begin{table*}[t]
\scriptsize
	\begin{center}
		\caption{
		Comparison of performance for RSCD on DSIFN-CD, CLCD and SYSU-CD datasets. Highest scores are in bold. All scores are in percentage.
		}
		\label{table:1}
            \begin{tabular}{l||ccccc||ccccc||ccccc}
		\Xhline{1.2pt}
            \rowcolor{mygray}
		     &\multicolumn{5}{c||}{DSIFN-CD} &\multicolumn{5}{c||}{CLCD} &\multicolumn{5}{c}{SYSU-CD}\\
            \rowcolor{mygray}
			\multicolumn{1}{c||}{\multirow{-2}{*}{Method}}
                &F1 &Pre. &Rec. &IoU &OA  &F1 &Pre. &Rec. &IoU &OA &F1 &Pre. &Rec. &IoU &OA \\			
                \hline \hline
                FC-EF \cite{fc-siam} &  59.71  & 61.80  &   57.75 &  42.56 & 86.77 &48.64 &73.34 &36.29 &32.14 &94.30 & 72.57 & 72.28& 72.85&56.94 & 87.01  \\
   FC-Siam-Di \cite{fc-siam}  &  62.95 & 68.44 & 58.27 & 45.93  & 88.35 &44.10 &72.97 &31.60 &28.29 &94.04 & 64.08& 84.94& 51.45& 47.15&86.40\\
   FC-Siam-Conc \cite{fc-siam} &  60.88  & 59.08 &  62.80 &  43.76 & 86.30 &54.48&68.21&45.22&37.35&94.35& 76.89& 83.03& 71.62& 62.45& 89.35\\
   IFNet \cite{ifnet} & 60.10 &  67.86 &   53.94 & 42.96 & 87.83 &48.65&49.96&47.41&32.14&92.55& 76.53& 79.59& 73.58& 61.91& 89.17\\
   DTCDSCN \cite{dtcdscn} &  63.72 & 53.87 & 77.99 & 46.76 & 84.91 &60.13&62.98&57.53&42.99&94.32& 80.11& 83.19& 77.25& 66.82& 90.96\\
   BIT \cite{bit}&  69.26 &  68.36 & 70.18 & 52.97 & 89.41 &67.10&73.07&62.04&50.49&95.47& 79.24& 84.89& 74.29 & 65.62& 90.82\\
   SNUNet \cite{snunet} & 66.18  &  60.60 & 72.89 & 49.45 &  87.34 &66.70&73.76 &60.88&50.04&95.48& 77.43& 81.93& 73.39& 63.17& 89.91\\
   ChangeStar \cite{changestar} &-  &  - &-  & - & - &60.75&62.23&59.34&43.63&94.3& -  &  - &-  & - & -\\
    DMATNet \cite{dmatnet}& 71.23  &  66.65 & 76.50 & 55.32 & 87.12 &66.56&72.74&61.34&49.87&95.41 & -  &  - &-  & - & -\\
   LGPNet \cite{lgpnet} & 63.20 & 49.96& \bf85.97 &46.19  & 82.99  &63.03&70.54&56.96&46.01&95.03 & 77.63&80.97   & 74.56 &63.44  &89.87 \\
   ChangeFormer \cite{changeformer}& 64.52 & 57.90 & 72.85 & 47.63 & 87.72 &58.44&65.00&53.07&41.28&94.38& 78.21& 79.37& 77.08& 64.22& 89.87\\
   SARASNet  \cite{sarasnet} & 67.49 & 68.18  & 66.81 & 50.93 & 88.96 & 74.70 & 76.68 & 72.83 & 59.62  & 96.33& 79.24& \bf86.35& 73.21& 65.62& 91.08\\
   			\hline
   Ours &  \bf74.75 & \bf75.96 & 73.57 & \bf59.68 & \bf91.55  & \bf78.55 & \bf79.87 & \bf77.27  & \bf64.67 &  \bf96.86	& \bf82.84 & 78.85 & \bf87.25 & \bf70.70 &	\bf91.47	\\
			\hline
		\end{tabular}
	\end{center}
\end{table*}

\setlength{\tabcolsep}{4pt}
\begin{table*}[t]
\scriptsize
	\begin{center}
		\caption{
		Comparison of performance for RSCD on LEVIR-CD and WHU-CD datasets. Highest scores are in bold. All scores are in percentage. Flops are counted with image size of $256 \times 256 \times 3$.
		}
		\label{table:2}
            \begin{tabular}{l||cc||ccccc||ccccc}
		\Xhline{1.2pt}
            \rowcolor{mygray}
		     & & &\multicolumn{5}{c||}{LEVIR-CD} &\multicolumn{5}{c}{WHU-CD}\\
            \rowcolor{mygray}
			\multicolumn{1}{c||}{\multirow{-2}{*}{Method}}
               & \multirow{-2}{*}{Params (M)}& \multirow{-2}{*}{Flops (G)} &F1 &Pre. &Rec. &IoU &OA  &F1 &Pre. &Rec. &IoU &OA \\			
                \hline \hline
                FC-EF \cite{fc-siam} &  1.10 & 1.55 &83.4 &86.91 &80.17 &71.53 &98.39 &72.01 &77.69 &67.10 &56.26 &92.07 \\
   FC-Siam-Di \cite{fc-siam}& 1.35 & 4.25  &86.31 &89.53 &83.31 &75.92 &98.67 &58.81 &47.33 &77.66 &41.66 &95.63 \\
   FC-Siam-Conc \cite{fc-siam}& 1.55 & 4.86 &83.69 &91.99 &76.77&71.96&98.49 &66.63 &60.88 &73.58 &49.95 &97.04  \\
   IFNet \cite{ifnet}& 50.71  &  41.18  &88.13 &\bf94.02 &82.93&78.77&98.87 &83.40 &\bf96.91 &73.19 &71.52 &98.83\\
   DTCDSCN \cite{dtcdscn}& 41.07  &  14.42 &87.67 &88.53&86.83&78.05&98.77 &71.95 &63.92 &82.30 &56.19 &97.42 \\
   BIT \cite{bit}& 11.89  & 8.71 &89.31 &89.24 &89.37&80.68&98.92 &83.98 &86.64 &81.48 &72.39 &98.75 \\
   SNUNet \cite{snunet}&  12.03  & 46.70 &88.16 &89.18 &87.17&78.83&98.82 &83.50 &85.60 &81.49 &71.67 &98.71  \\
   ChangeStar \cite{changestar}& 16.77 & 68.52 &89.30 &89.88 &88.72&80.66&98.90 &87.01 &88.78 &85.31 &77.00 &98.70\\
   DMATNet \cite{dmatnet}&  13.27 & - & 89.97 &90.78 &89.17&81.83 & 98.06 & 85.07 & 89.46 & 82.24 & 74.98 & 95.83\\ 
   LGPNet \cite{lgpnet}& 70.99  &  125.79 &89.37 &93.07 &85.95&80.78&99.00 &79.75 &89.68 &71.81 &66.33 &98.33\\
   ChangeFormer \cite{changeformer}& 41.03 & 202.86 &90.40 &92.05&88.80&82.48&99.04 &81.82 &87.25 &77.03 &69.24 &94.80\\
   SARASNet \cite{sarasnet}& 56.89 & 139.9 &  90.44   &91.42  & \bf89.48 &82.55  & \bf99.11  & 89.55 & 88.68 & 90.44  & 81.08 & 99.05  \\
   			\hline
   Ours & 24.49 & 32.46 & \bf90.66 & 92.01  &  89.35 & \bf82.92 & 99.06   & \bf91.56 & 92.25 & \bf90.89 & \bf84.44 & \bf99.23 \\
			\hline
		\end{tabular}
	\end{center}
\end{table*}

\section{Experiments}
\label{sec:main_exp}
We perform experiments on the DSIFN-CD \cite{dsifn}, LEVIR-CD \cite{levir}, CLCD \cite{clcd}, SYSU-CD \cite{sysucd} and WHU-CD \cite{whu} datasets. We implement CDMaskFormer using PyTorch. The initial learning rate is 0.0001 (0.0005 for backbone) and the Adam optimizer is adopted with a weight decay of 0.05 (0.01 for backbone). \textit{More details and experimental results are provided in the Appendix.}

\setlength{\tabcolsep}{3pt}
\begin{table*}[t]
\scriptsize
 \begin{minipage}{0.5\linewidth}
    \vspace{0pt}
    \centering
		\caption{
		Ablation experiments with different change extractors on the DSIFN-CD dataset.
		}
  \vspace{-1mm}
            \label{table:3}
            \begin{tabular}{c||cc||ccccc}
		\Xhline{1.2pt}
            \rowcolor{mygray}
		     Extractor &Params &Flops&F1 &Pre. &Rec. &IoU &OA\\		
                \hline \hline
                CNN-CE & 32.73 & 12.81 & 71.18 & 75.26 & 67.52 & 55.26 & 90.71\\
                TR1-CE & 56.82 & 21.70  & 72.44 & 72.89 & 71.99 & 56.78 & 90.83 \\
                TR2-CE & 35.26 & 16.85 & 71.67 & 74.30 & 69.22 & 55.85 & 90.70\\
                Ours-CE & 33.52  & 13.12  & \bf74.75 & \bf75.96 & \bf73.57 & \bf59.68 & \bf91.55 \\
   			\hline
		\end{tabular}
 \end{minipage}
\begin{minipage}{0.48\linewidth}
    \vspace{0pt}
	\centering
		\caption{
		Ablation experiments for varying the structure of our change extractor on the DSIFN-CD dataset.
		}
  \vspace{-1mm}
		\label{table:4}
            \begin{tabular}{cc||ccccc}
		\Xhline{1.2pt}
            \rowcolor{mygray}
		     ${W'}_i^t$ & ${W'}_i^c$ &F1 &Pre. &Rec. &IoU &OA\\		
                \hline \hline
               \CheckmarkBold & \CheckmarkBold & \bf74.75 & \bf75.96 & \bf73.57 & \bf59.68 & \bf91.55 \\
                 \CheckmarkBold & \XSolidBrush  & 71.48 & 74.82 & 68.43 & 55.62  & 90.72  \\
                \XSolidBrush  &  \CheckmarkBold & 73.54 & 75.55 & 71.64 & 58.16 & 91.24 \\
                 \XSolidBrush & \XSolidBrush  & 70.39  & 70.82  & 69.96 & 53.50 & 90.00  \\
   			\hline
		\end{tabular}
 \end{minipage}
\end{table*}

\setlength{\tabcolsep}{5pt}
\begin{table*}[t]
\scriptsize
\begin{minipage}{0.52\linewidth}
    \vspace{0pt}
	\centering
		\caption{
		Ablation experiments for the structure of the detail-enhanced transformer decoder.
		}
  \vspace{-1mm}
		\label{table:5}
            \begin{tabular}{c||ccccc}
		\Xhline{1.2pt}
            \rowcolor{mygray}
		      &F1 &Pre. &Rec. &IoU &OA\\		
                \hline \hline
                MAB $\times$ 3 & 70.84 & 69.15 & 72.63 & 54.85 & 89.88 \\
                MAB $\times$ 4 & 72.71 & 72.42 & 73.00 & 57.12 & 90.69\\
                MAB $\times$ 3 + SAAB-v1 & 71.58 & 71.95 & 71.22 & 55.74 & 90.39 \\
                MAB $\times$ 3 + SAAB & \bf74.75 & \bf75.96 & \bf73.57 & \bf59.68 & \bf91.55\\
   			\hline
		\end{tabular}
\end{minipage}
\begin{minipage}{0.05\linewidth}
\vspace{0pt}
	\centering
 \end{minipage}
\begin{minipage}{0.45\linewidth}
\vspace{0pt}
	\centering
    \caption{
        Ablation experiments of different change query numbers.
		}
  \vspace{-1mm}
		\label{table:6}
            \begin{tabular}{c||ccccc}
		\Xhline{1.2pt}
            \rowcolor{mygray}
		      $m$ &F1 &Pre. &Rec. &IoU &OA\\		
                \hline \hline
                3 & 72.13 & 66.23 & 79.18  & 56.41 & 89.60 \\
                5  & \bf74.75 & \bf75.96 & 73.57 & \bf59.68 & \bf91.55   \\
                10 & 72.19 & 65.35 & \bf80.62 & 56.48 & 89.45   \\
                20  & 73.28  &  71.36  & 75.32  & 57.83 &  90.67 \\
   			\hline
		\end{tabular}
\end{minipage}
\end{table*}

\subsection{Main Results}

The experimental results are shown in Tables~\ref{table:1} and~\ref{table:2}. It can be observed that the proposed CDMaskFormer achieves state-of-the-art performance on the four change detection datasets. Specifically, CDMaskFormer achieves 2.01\%, 7.26\%, 3.60\% and 3.85\% F1 metrics enhancement on WHU-CD, DSIFN-CD, SYSU-CD and CLCD, respectively, compared to the recent method SARASNet. On the LEVIR-CD dataset, CDMaskFormer boosted less, only 0.22\%. This can be hypothesized as CDMaskFormer can improve more significantly in the case of more complex scenarios, more diverse object distributions, and richer variations (e.g., DSIFN-CD and CLCD dataset). In addition, CDMaskFormer only has 33.52M parameters and 13.12G Flops, which requires only 58.9\% of parameter requirements and 9.4\% of computational consumption compared to SARASNet. Compared to other models such as BIT and ChangeFormer, CDMaskFormer improves more significantly and has a better balance between performance and efficiency.

\subsection{Ablation Studies}
\label{ablation}

\textbf{Structure design of change extractor}
The two key components of the change extractor are the weights $\mathcal{W}_i^t$ on the space and $\mathcal{W}_i^c$ on the channel. Therefore, we experimentally explore their validity as shown in Table~\ref{table:4}. When both $\mathcal{W}_i^t$ and $\mathcal{W}_i^c$ are absent, i.e., we obtain the change representations by direct subtraction, the model achieves 70.39\% F1. After introducing the change extractor, the results show that both $\mathcal{W}_i^t$ and $\mathcal{W}_i^c$ contribute to the performance of the model and the model reaches the highest 74.75\% F1 when they are combined. We observe that $\mathcal{W}_i^c$ is more conductive to improving the performance, which might be because treating all channels equally brings a lot of redundant information. And the introduction of $\mathcal{W}_i^c$ can reduce the redundancy of channels and lead the model to focus on significant channels adaptively. In addition, we also compare some classical change extractors (CNN-CE \cite{base_cnn}, TR1-CE \cite{sarasnet}, TR2-CE \cite{base_tr1}) as shown in the Table \ref{table:3}. The results validate that the designed change extractors are able to obtain high-quality change representations at a small computational cost.

\textbf{Structure design of detail-enhanced decoder.}
We focus on exploring the effects of the SAAB and scene context $G$ on the experimental results, as shown in Table~\ref{table:5}. When SAAB is not present, the model only obtains 70.84\% of the F1 value. And when SAAB is present and built based on ordinary axial attention (i.e., SAAB-v1), there is a 0.74\% improvement in the model performance. It might be because the introduction of high-resolution feature maps drives the model to mine more details of the changed areas. Furthermore, the model achieves the best performance when scene context-guided axial attention is used to construct the SAAB, which might make up the lack of scene context introduced by the axial attention. The experimental results validate the effectiveness of the SAAB structure design.

\textbf{Number of change queries.}
We explore the effect of the number of change queries $m$ on the model performance as shown in Table~\ref{table:6}. It can be observed that the model achieves optimal results when $m$ is 5. Moreover, when $m$ decreases or increases, the results are reduced. It might be because fewer change queries lead to reduced fault tolerance, whereas a greater number of change queries introduces more complexity. Therefore, we choose $m$ to be 5 in the final model.

\setlength{\tabcolsep}{3pt}
\begin{wraptable}{r}{0.5\linewidth}
\scriptsize
	\centering
		\caption{
		Ablation experiments with coefficients of different loss functions on the DSIFN-CD dataset.
		}
  \vspace{-5pt}
		\label{table:7}
            \begin{tabular}{ccc||ccccc}
		\Xhline{1.2pt}
            \rowcolor{mygray}
		      $\lambda_{cls}$ & $\lambda_{dice}$ & $\lambda_{ce}$ &F1 &Pre. &Rec. &IoU &OA\\	
                \hline \hline
                 1 & 1& 1& 72.71 & 72.42 & 73.00 & 57.12 & 90.69 \\
                 2  &5 &  5 & 71.13 & 72.87 & 69.48 & 55.20  & 90.58 \\
                 1  & 10 & 10 & \bf74.75 & \bf75.96 & \bf73.57 & \bf59.68 & \bf91.55  \\
                 2  & 1 & 1 & 69.44  &  70.94  & 67.99  &  53.18  &  89.83 \\
   			\hline
		\end{tabular}
  \vspace{-10pt}
\end{wraptable}

\textbf{Coefficients of different loss functions.}
We also explore the effect of different coefficients on the results as shown in Table~\ref{table:7}. We select some commonly used coefficients. It is experimentally verified that CDMaskFormer achieves the best performance at $\lambda_{cls}=1$, $\lambda_{dice}=10$, and $\lambda_{ce}=10$. It might be because the class classification task is relatively easier with only a few prototypes and two classes, resulting in the model not requiring as much attention to the classification loss $\lambda_{cls}$.

\section{Conclusion}
In this paper, we first analyze the inadequacy of the existing pixel-by-pixel change detection paradigm, i.e., the fixed change prototype cannot tolerate the richness of changes due to complex scene and imaging condition variations. To address the challenge, we rethink the change detection task from a mask view and further propose the corresponding 1) meta-architecture CDMask and 2) instance network CDMaskFormer. the CDMask consists of a Siamese backbone, a change extractor, a pixel decoder, a transformer decoder, and a normalized detector. In particular, the normalized detector is proposed to ensure the proper functioning of CDMask. To further exploit the potential of CDMask, CDMaskFormer is proposed, whose components are highly customized for the change detection task, aiming at trying to get a better performance with a more lightweight operation. Numerous experiments have validated the effectiveness of CDMask and CDMaskFormer. In the future we will explore more effective component design of CDMask and try to build a unified architecture for remote sensing image segmentation and change detection.

{
    \bibliographystyle{plain}
    \bibliography{ref}
}

\clearpage
\appendix
\centerline{\maketitle{\textbf{APPENDIX}}}
In the appendix, we provide the following items that shed deeper insight on our contributions:
\begin{itemize}
    \item  \S\ref{sec:exp}: More experimental results.
    \item \S\ref{sec:detail}: Dataset and Implementation Details.
    \item  \S\ref{sec:vis}: More qualitative visualization.
    \item  \S\ref{sec:cdpixel}: CDPixel analysis.
    \item  \S\ref{sec:limit}: Limitation analysis.
\end{itemize}

\setlength{\tabcolsep}{6pt}
\begin{table*}[h]
	\begin{center}
		\caption{
        Ablation experiments of different threshold $\mathcal{T}$ on DSIFN-CD and CLCD datasets.
		}
		\label{table:t}
            \begin{tabular}{c||ccccc||ccccc}
		\Xhline{1.2pt}
            \rowcolor{mygray}
		     &\multicolumn{5}{c||}{DSIFN-CD} &\multicolumn{5}{c}{CLCD}\\
            \rowcolor{mygray}
			\multicolumn{1}{c||}{\multirow{-2}{*}{$\mathcal{T}$}}
                &F1 &Pre. &Rec. &IoU &OA  &F1 &Pre. &Rec. &IoU &OA \\			
                \hline \hline
                0.2 &  60.48 &  45.75 &  \bf89.19  & 43.35  &  80.19 & 76.67 & 76.50 & 76.83  & 62.16 & 96.52 \\
                0.3 & 69.92  & 59.36  & 85.04   & 53.75  &  87.57 & 76.83  & 78.74 & 75.01 & 62.37 & 96.63 \\
                0.4 & 74.57  & 68.72  & 81.51   &  59.45 &  90.55 & 77.23  & 76.00 & \bf78.51 & 62.91 & 96.56 \\
                0.5 & \bf74.75  & 75.96  &  73.57  & \bf59.68  &  \bf91.55 & \bf78.55 & 79.87 & 77.27 & \bf64.67 & 96.86 \\
                0.6 & 72.44  & 72.89  &  71.99  &  56.78 & 90.83  &  78.36 &  81.71 &  75.27  & 64.42  &  96.91  \\
                0.7 & 74.02  & 74.53  &  73.52  & 58.76  &  91.25  & 77.83  & 82.87  & 73.37   & 63.71  &  96.89 \\
                0.8 &  73.38 &  \bf76.45 & 70.55   &  57.96 & 91.20   &  77.65 & \bf84.30  &  71.97  & 63.47  &  \bf96.92 \\
   			\hline
		\end{tabular}
	\end{center}
\end{table*}

\begin{figure}[h]
\centering
\captionsetup[subfloat]{labelsep=none,format=plain,labelformat=empty,font=tiny}
\subfloat[$T_1$]{
\begin{minipage}[t]{0.12\linewidth}
\includegraphics[width=1\linewidth]{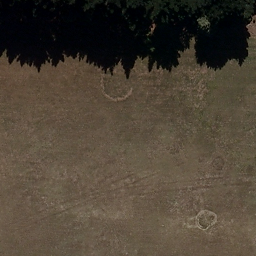}\vspace{2pt}
\includegraphics[width=1\linewidth]{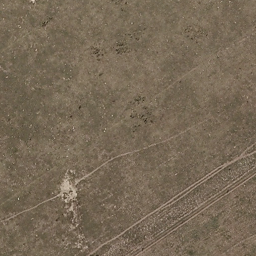}\vspace{2pt}
\includegraphics[width=1\linewidth]{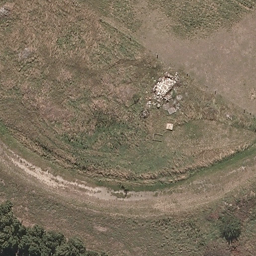}\vspace{2pt}
\includegraphics[width=1\linewidth]{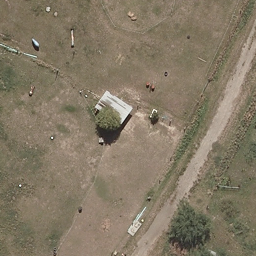}\vspace{2pt}
\end{minipage}}
\subfloat[$T_2$]{
\begin{minipage}[t]{0.12\linewidth}
\includegraphics[width=1\linewidth]{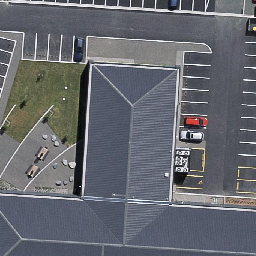}\vspace{2pt}
\includegraphics[width=1\linewidth]{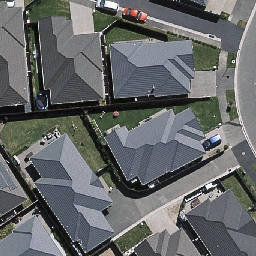}\vspace{2pt}
\includegraphics[width=1\linewidth]{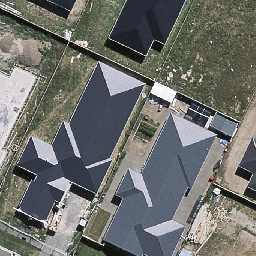}\vspace{2pt}
\includegraphics[width=1\linewidth]{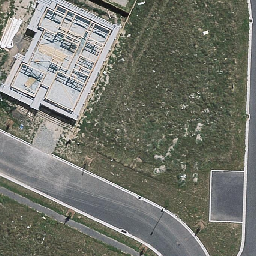}\vspace{2pt}
\end{minipage}}
\subfloat[GT]{
\begin{minipage}[t]{0.12\linewidth}
\includegraphics[width=1\linewidth]{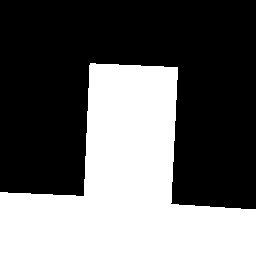}\vspace{2pt}
\includegraphics[width=1\linewidth]{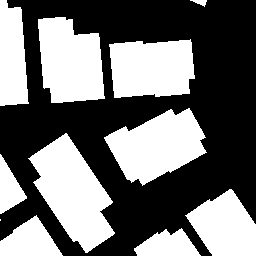}\vspace{2pt}
\includegraphics[width=1\linewidth]{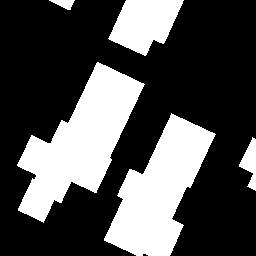}\vspace{2pt}
\includegraphics[width=1\linewidth]{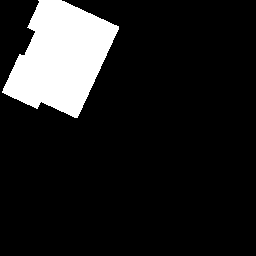}\vspace{2pt}
\end{minipage}}
\subfloat[SNUNet]{
\begin{minipage}[t]{0.12\linewidth}
\includegraphics[width=1\linewidth]{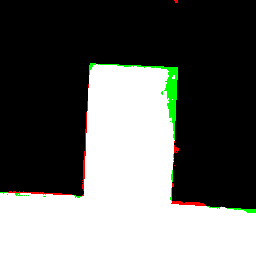}\vspace{2pt}
\includegraphics[width=1\linewidth]{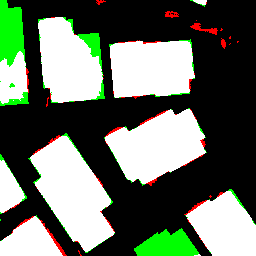}\vspace{2pt}
\includegraphics[width=1\linewidth]{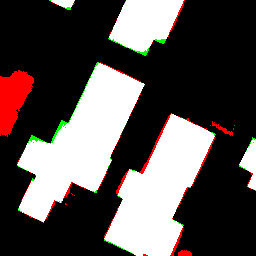}\vspace{2pt}
\includegraphics[width=1\linewidth]{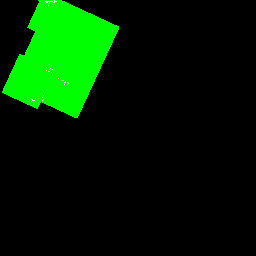}\vspace{2pt}
\end{minipage}}
\subfloat[BIT]{
\begin{minipage}[t]{0.12\linewidth}
\includegraphics[width=1\linewidth]{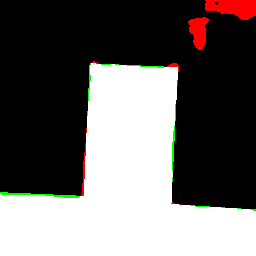}\vspace{2pt}
\includegraphics[width=1\linewidth]{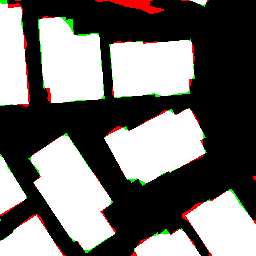}\vspace{2pt}
\includegraphics[width=1\linewidth]{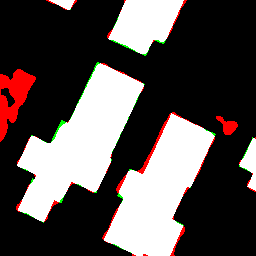}\vspace{2pt}
\includegraphics[width=1\linewidth]{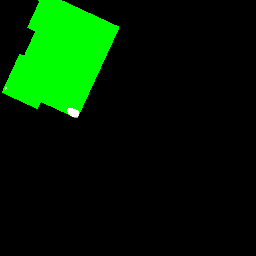}\vspace{2pt}
\end{minipage}}
\subfloat[SARASNet]{
\begin{minipage}[t]{0.12\linewidth}
\includegraphics[width=1\linewidth]{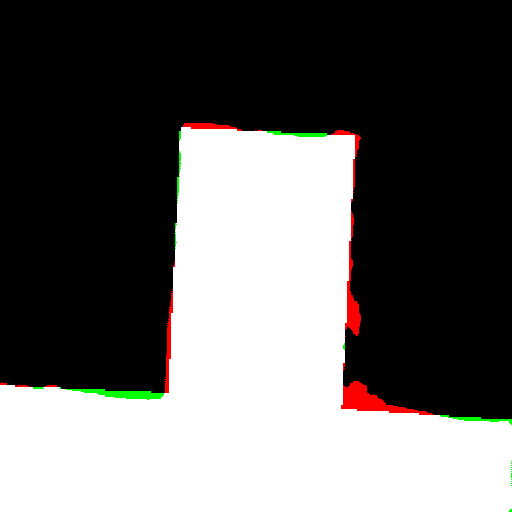}\vspace{2pt}
\includegraphics[width=1\linewidth]{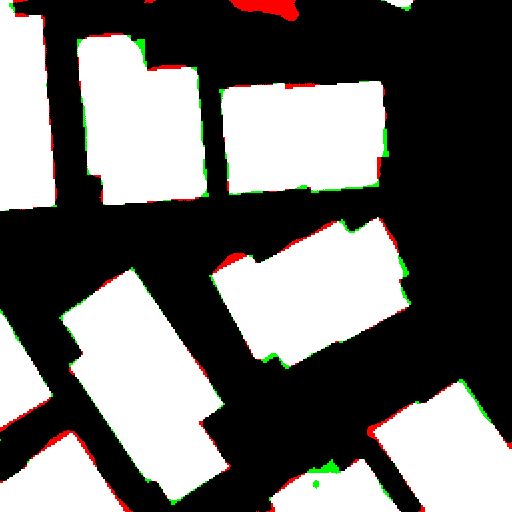}\vspace{2pt}
\includegraphics[width=1\linewidth]{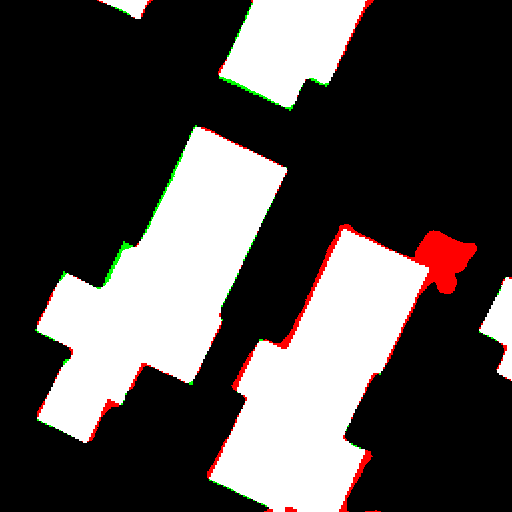}\vspace{2pt}
\includegraphics[width=1\linewidth]{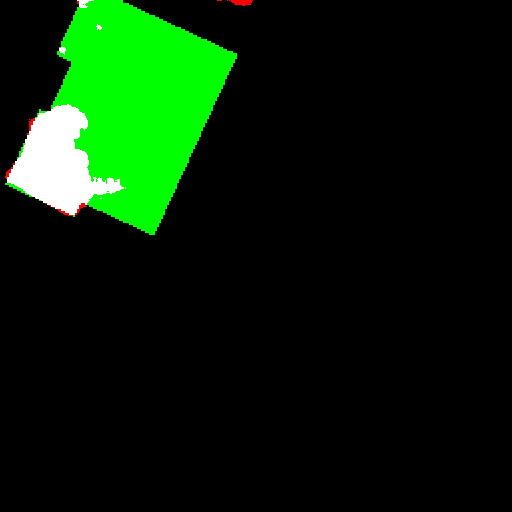}\vspace{2pt}
\end{minipage}}
\subfloat[CDMaskFormer]{
\begin{minipage}[t]{0.12\linewidth}
\includegraphics[width=1\linewidth]{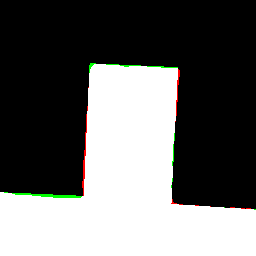}\vspace{2pt}
\includegraphics[width=1\linewidth]{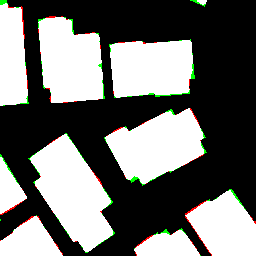}\vspace{2pt}
\includegraphics[width=1\linewidth]{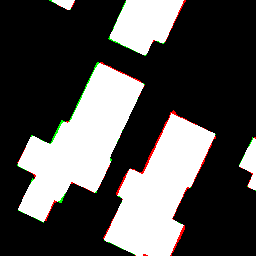}\vspace{2pt}
\includegraphics[width=1\linewidth]{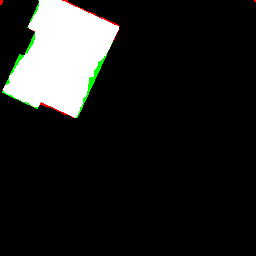}\vspace{2pt}
\end{minipage}}
\caption{Example results output from RSCD methods on test sets from WHU-CD dataset. Pixels are colored differently for better visualization (i.e., white for true positive, black for true negative, red for false positive, and green for false negative).}
\label{fig:whu}
\end{figure}
\section{More experimental results.}\label{sec:exp}
We perform ablation analysis of the threshold $\mathcal{T}$ in Table \ref{table:t}, where we set $\mathcal{T}$ to 0.2, 0.3, 0.4, 0.5, 0.6, 0.7, and 0.8, respectively. The experimental results show that the model achieves optimal performance when $\mathcal{T}$ is 0.5, i.e., it achieves F1 values of 74.75\% and 78.55\% on DSIFN and CLCD datasets, respectively. The experimental results show a trend that Precision increases and Recall decreases when the threshold $\mathcal{T}$ is increased and vice versa. 

\begin{figure}[t]
\centering
\captionsetup[subfloat]{labelsep=none,format=plain,labelformat=empty,font=tiny}
\subfloat[$T_1$]{
\begin{minipage}[t]{0.12\linewidth}
\includegraphics[width=1\linewidth]{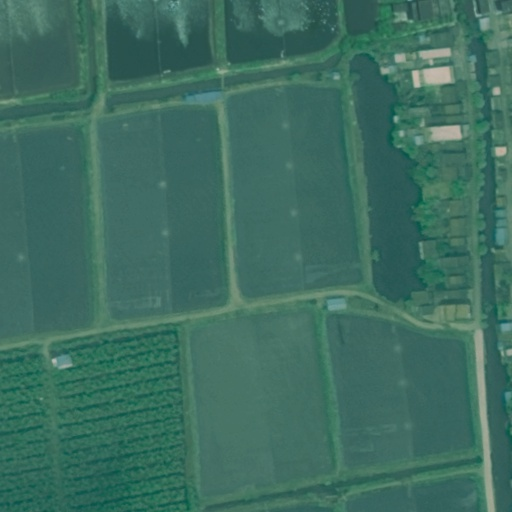}\vspace{2pt}
\includegraphics[width=1\linewidth]{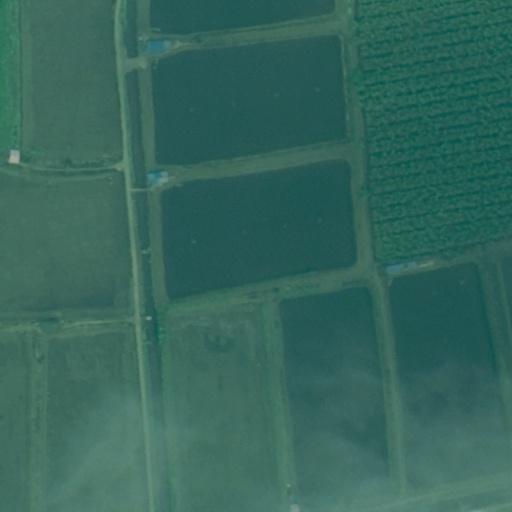}\vspace{2pt}
\includegraphics[width=1\linewidth]{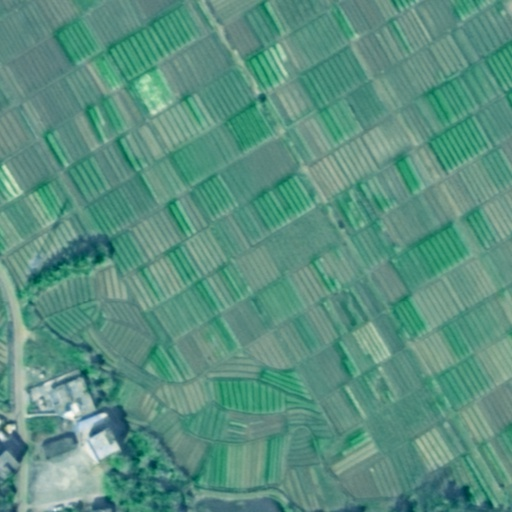}\vspace{2pt}
\includegraphics[width=1\linewidth]{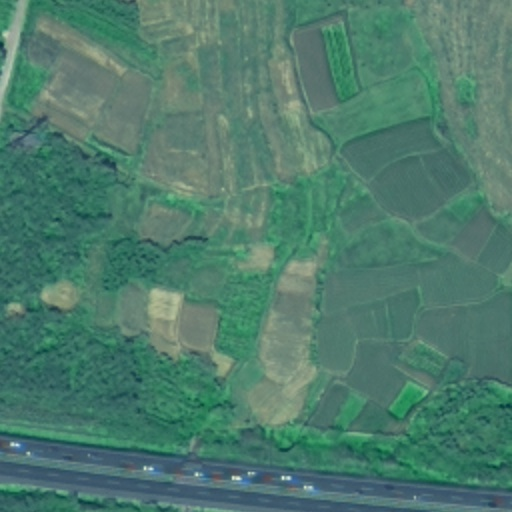}\vspace{2pt}
\end{minipage}}
\subfloat[$T_2$]{
\begin{minipage}[t]{0.12\linewidth}
\includegraphics[width=1\linewidth]{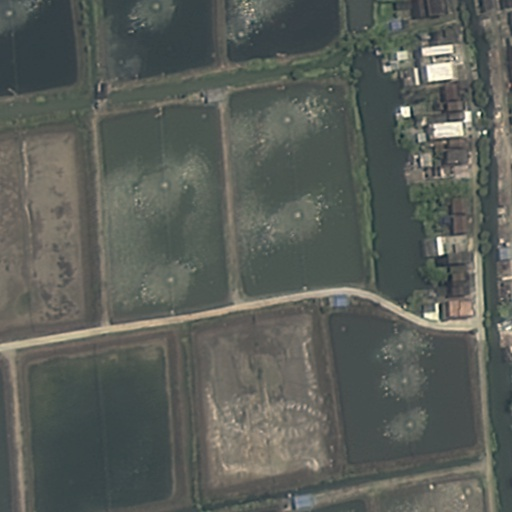}\vspace{2pt}
\includegraphics[width=1\linewidth]{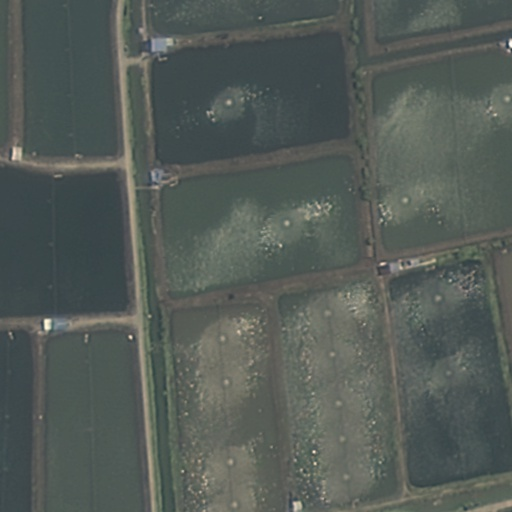}\vspace{2pt}
\includegraphics[width=1\linewidth]{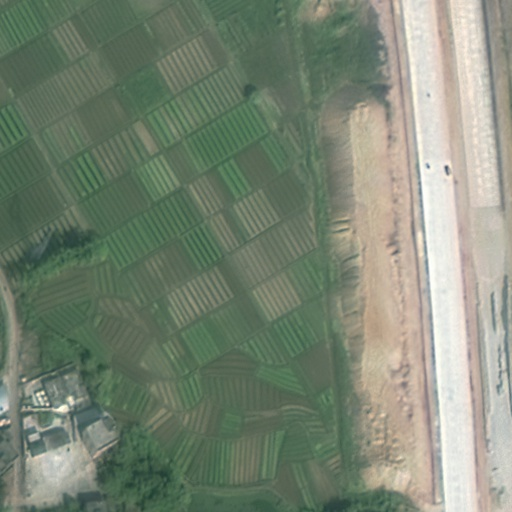}\vspace{2pt}
\includegraphics[width=1\linewidth]{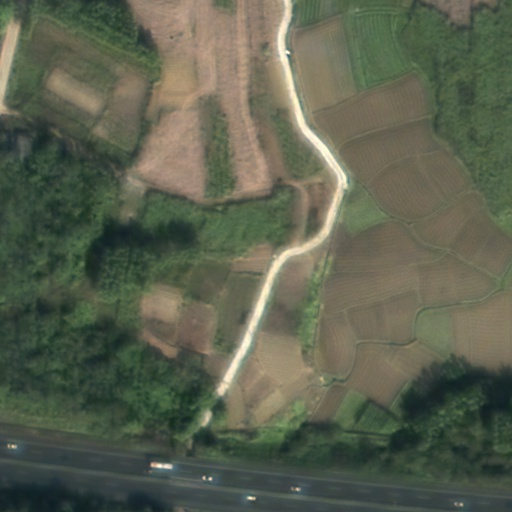}\vspace{2pt}
\end{minipage}}
\subfloat[GT]{
\begin{minipage}[t]{0.12\linewidth}
\includegraphics[width=1\linewidth]{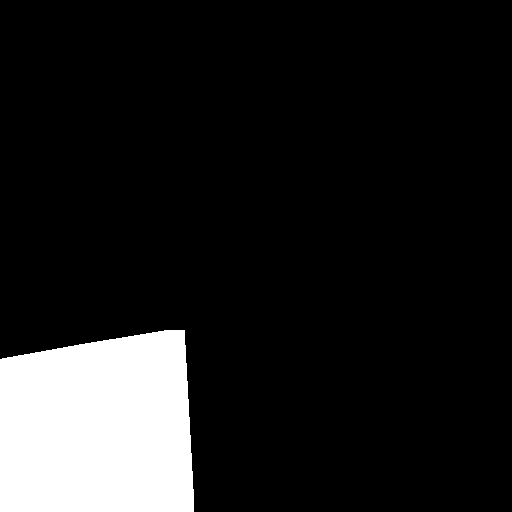}\vspace{2pt}
\includegraphics[width=1\linewidth]{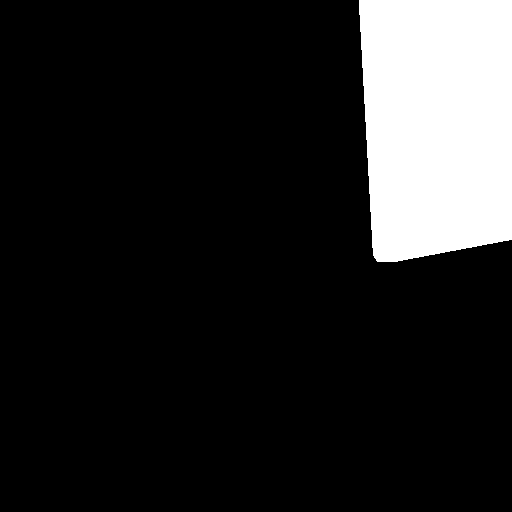}\vspace{2pt}
\includegraphics[width=1\linewidth]{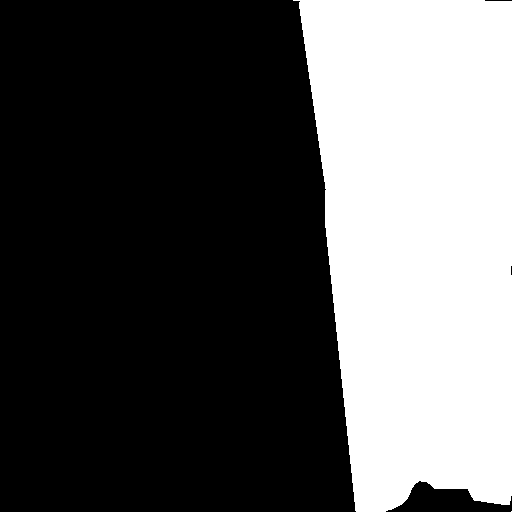}\vspace{2pt}
\includegraphics[width=1\linewidth]{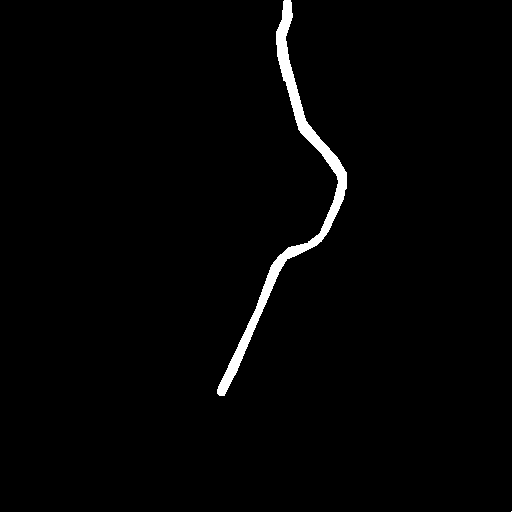}\vspace{2pt}
\end{minipage}}
\subfloat[SNUNet]{
\begin{minipage}[t]{0.12\linewidth}
\includegraphics[width=1\linewidth]{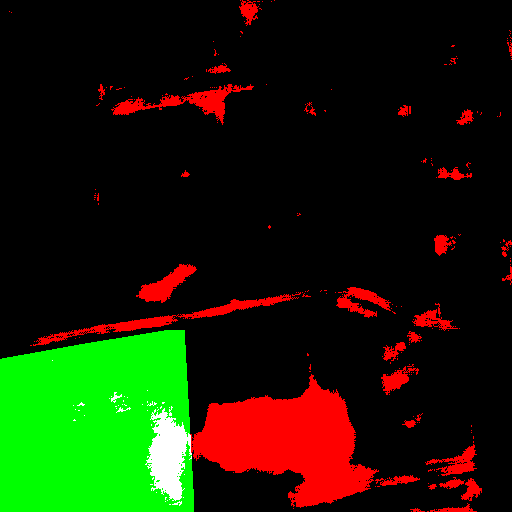}\vspace{2pt}
\includegraphics[width=1\linewidth]{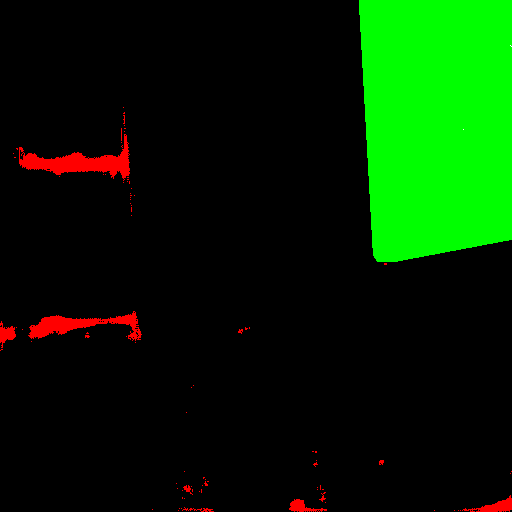}\vspace{2pt}
\includegraphics[width=1\linewidth]{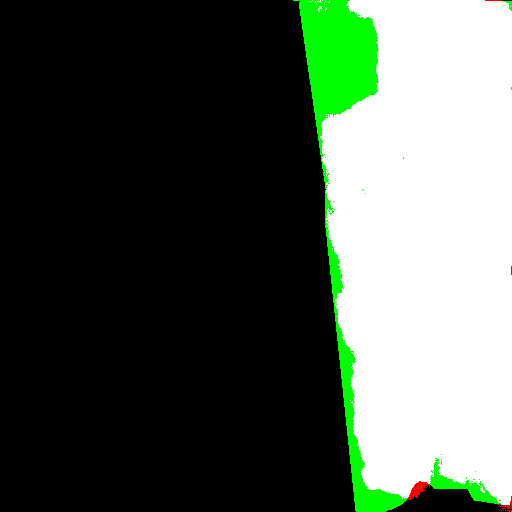}\vspace{2pt}
\includegraphics[width=1\linewidth]{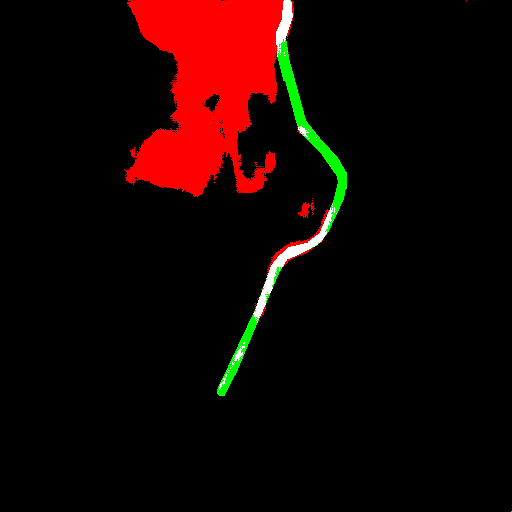}\vspace{2pt}
\end{minipage}}
\subfloat[BIT]{
\begin{minipage}[t]{0.12\linewidth}
\includegraphics[width=1\linewidth]{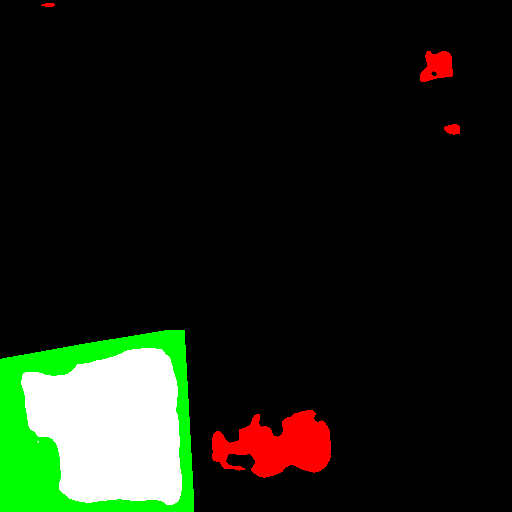}\vspace{2pt}
\includegraphics[width=1\linewidth]{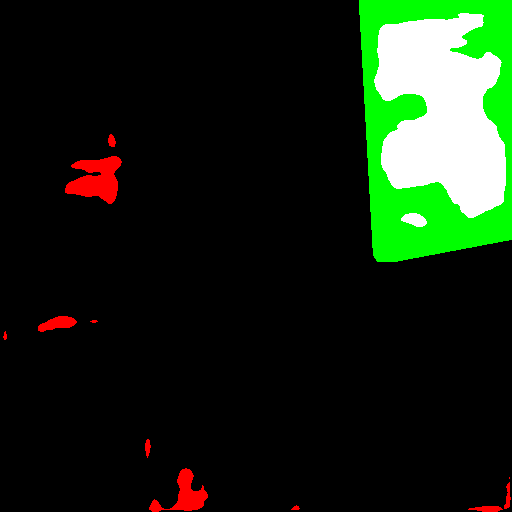}\vspace{2pt}
\includegraphics[width=1\linewidth]{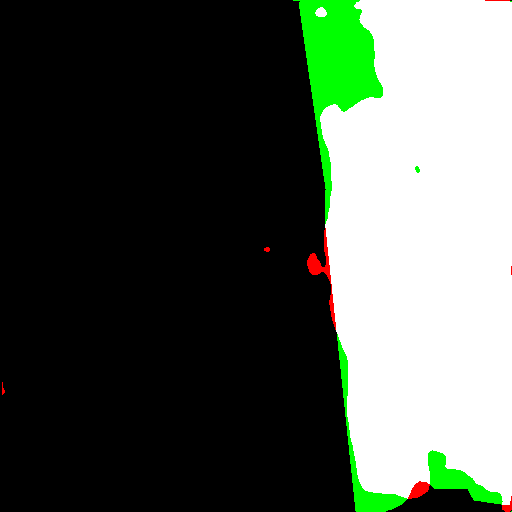}\vspace{2pt}
\includegraphics[width=1\linewidth]{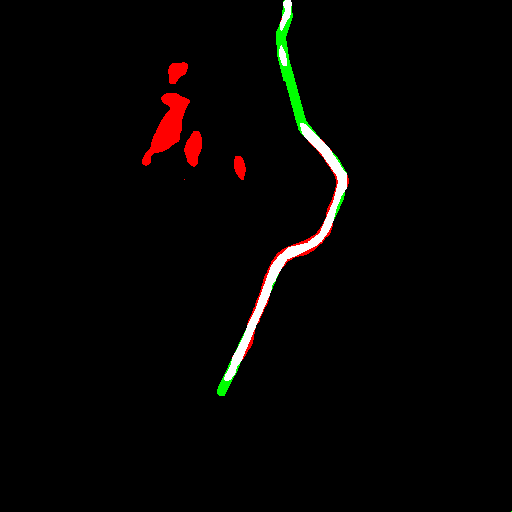}\vspace{2pt}
\end{minipage}}
\subfloat[SARASNet]{
\begin{minipage}[t]{0.12\linewidth}
\includegraphics[width=1\linewidth]{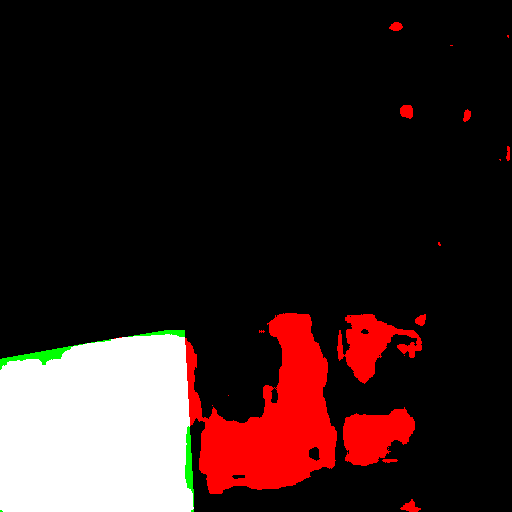}\vspace{2pt}
\includegraphics[width=1\linewidth]{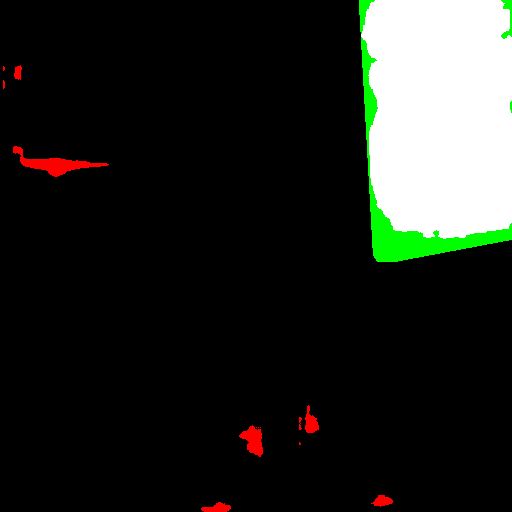}\vspace{2pt}
\includegraphics[width=1\linewidth]{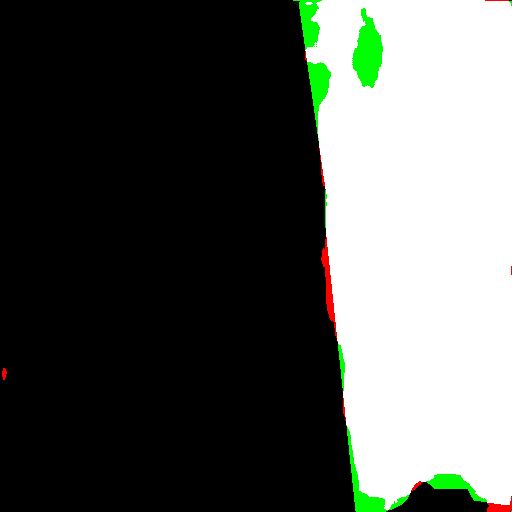}\vspace{2pt}
\includegraphics[width=1\linewidth]{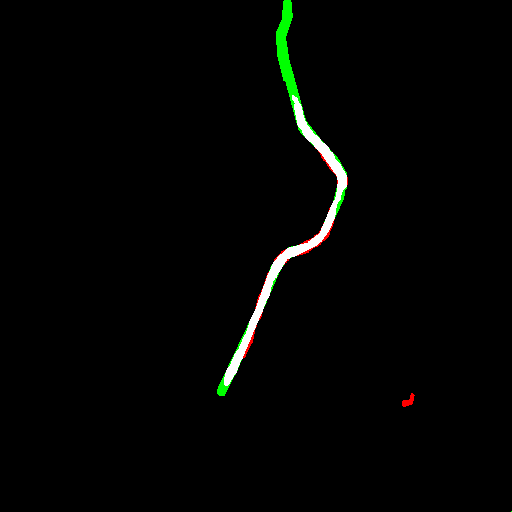}\vspace{2pt}
\end{minipage}}
\subfloat[CDMaskFormer]{
\begin{minipage}[t]{0.12\linewidth}
\includegraphics[width=1\linewidth]{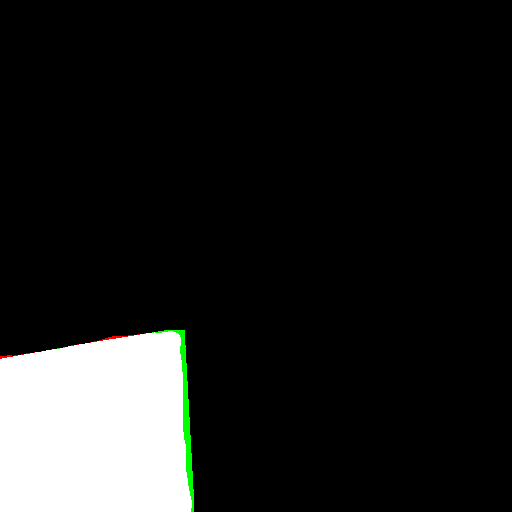}\vspace{2pt}
\includegraphics[width=1\linewidth]{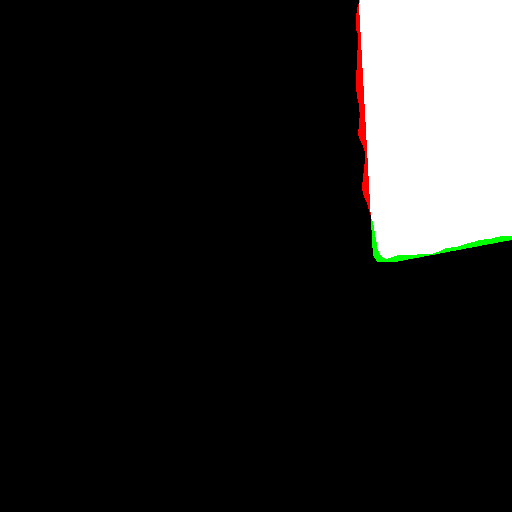}\vspace{2pt}
\includegraphics[width=1\linewidth]{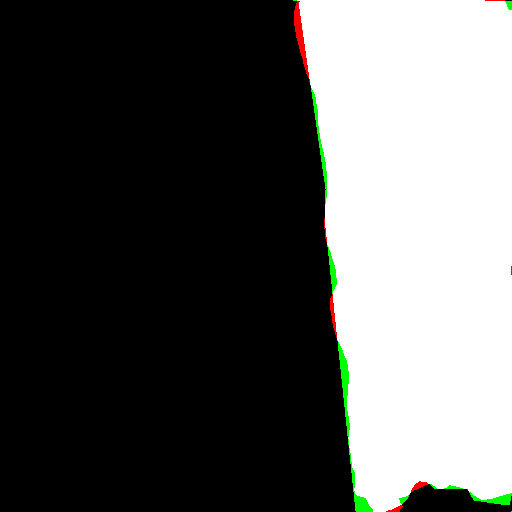}\vspace{2pt}
\includegraphics[width=1\linewidth]{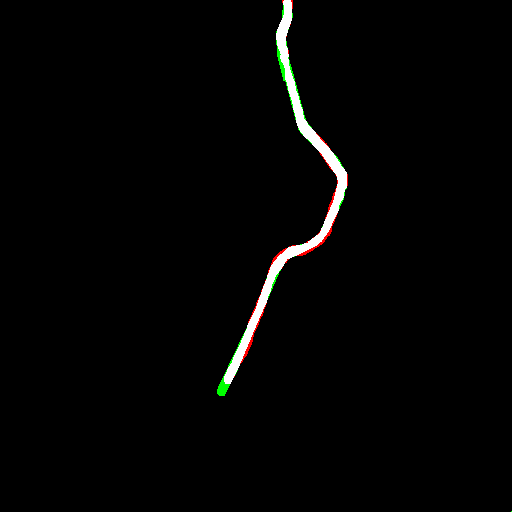}\vspace{2pt}
\end{minipage}}
\caption{Example results output from RSCD methods on test sets from CLCD dataset. Pixels are colored differently for better visualization (i.e., white for true positive, black for true negative, red for false positive, and green for false negative).}
\label{fig:clcd}
\end{figure}

\begin{figure}[t]
\centering
\captionsetup[subfloat]{labelsep=none,format=plain,labelformat=empty,font=tiny}
\subfloat[$T_1$]{
\begin{minipage}[t]{0.12\linewidth}
\includegraphics[width=1\linewidth]{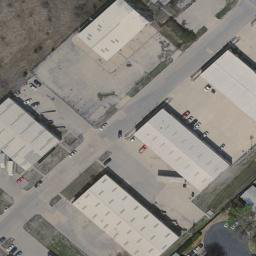}\vspace{2pt}
\includegraphics[width=1\linewidth]{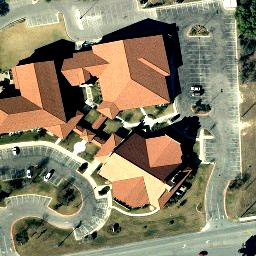}\vspace{2pt}
\includegraphics[width=1\linewidth]{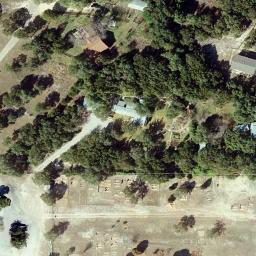}\vspace{2pt}
\includegraphics[width=1\linewidth]{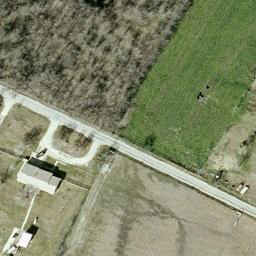}\vspace{2pt}
\end{minipage}}
\subfloat[$T_2$]{
\begin{minipage}[t]{0.12\linewidth}
\includegraphics[width=1\linewidth]{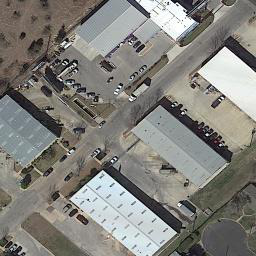}\vspace{2pt}
\includegraphics[width=1\linewidth]{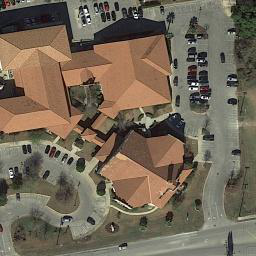}\vspace{2pt}
\includegraphics[width=1\linewidth]{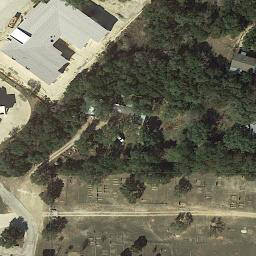}\vspace{2pt}
\includegraphics[width=1\linewidth]{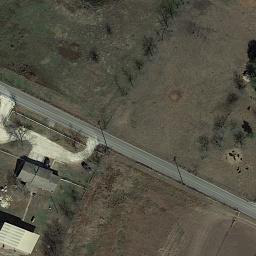}\vspace{2pt}
\end{minipage}}
\subfloat[GT]{
\begin{minipage}[t]{0.12\linewidth}
\includegraphics[width=1\linewidth]{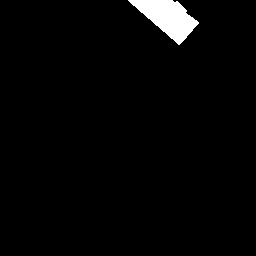}\vspace{2pt}
\includegraphics[width=1\linewidth]{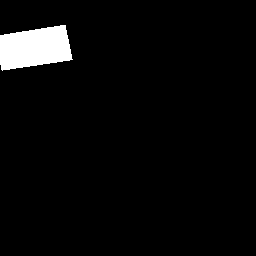}\vspace{2pt}
\includegraphics[width=1\linewidth]{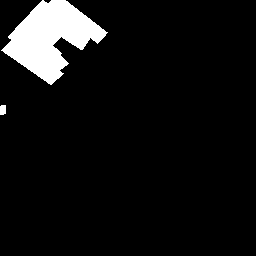}\vspace{2pt}
\includegraphics[width=1\linewidth]{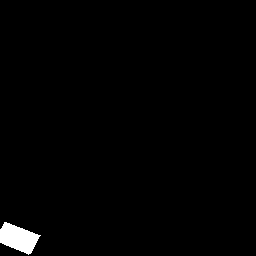}\vspace{2pt}
\end{minipage}}
\subfloat[SNUNet]{
\begin{minipage}[t]{0.12\linewidth}
\includegraphics[width=1\linewidth]{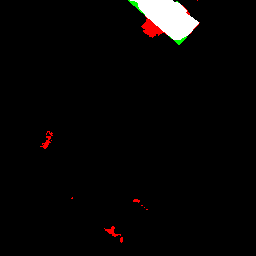}\vspace{2pt}
\includegraphics[width=1\linewidth]{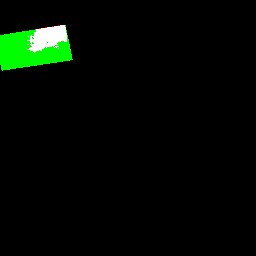}\vspace{2pt}
\includegraphics[width=1\linewidth]{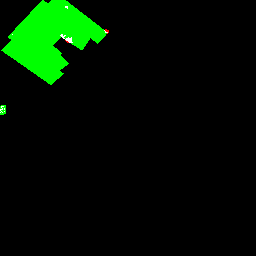}\vspace{2pt}
\includegraphics[width=1\linewidth]{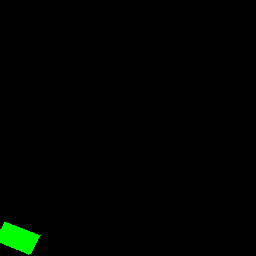}\vspace{2pt}
\end{minipage}}
\subfloat[BIT]{
\begin{minipage}[t]{0.12\linewidth}
\includegraphics[width=1\linewidth]{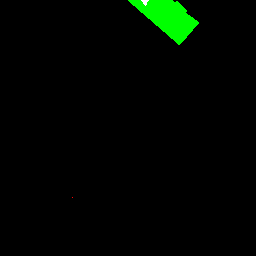}\vspace{2pt}
\includegraphics[width=1\linewidth]{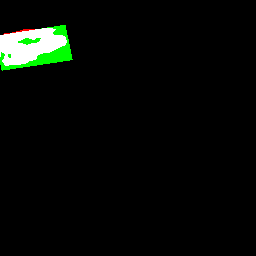}\vspace{2pt}
\includegraphics[width=1\linewidth]{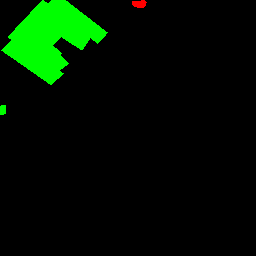}\vspace{2pt}
\includegraphics[width=1\linewidth]{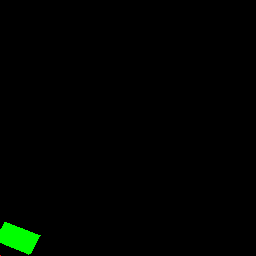}\vspace{2pt}
\end{minipage}}
\subfloat[SARASNet]{
\begin{minipage}[t]{0.12\linewidth}
\includegraphics[width=1\linewidth]{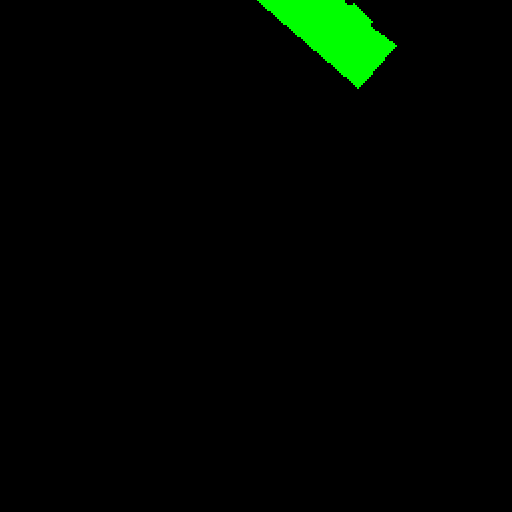}\vspace{2pt}
\includegraphics[width=1\linewidth]{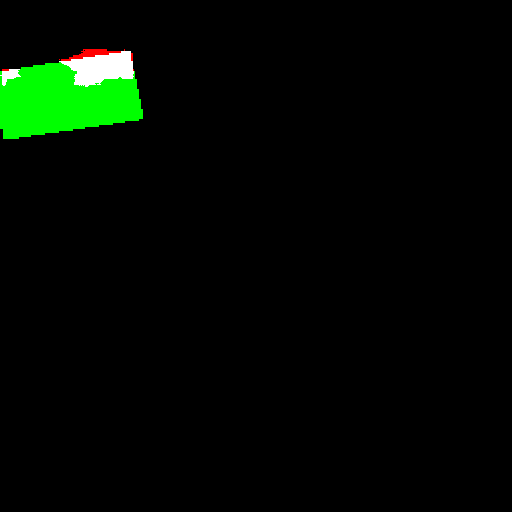}\vspace{2pt}
\includegraphics[width=1\linewidth]{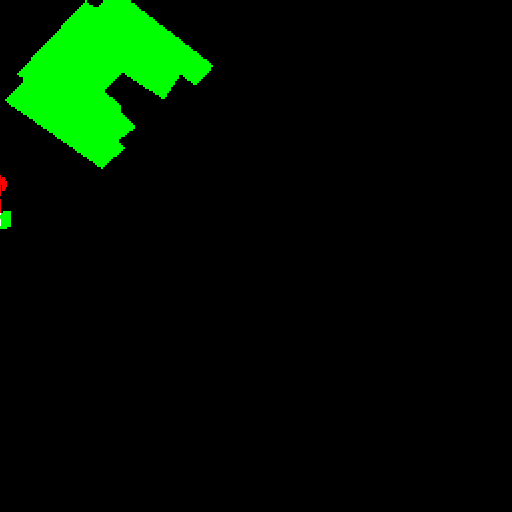}\vspace{2pt}
\includegraphics[width=1\linewidth]{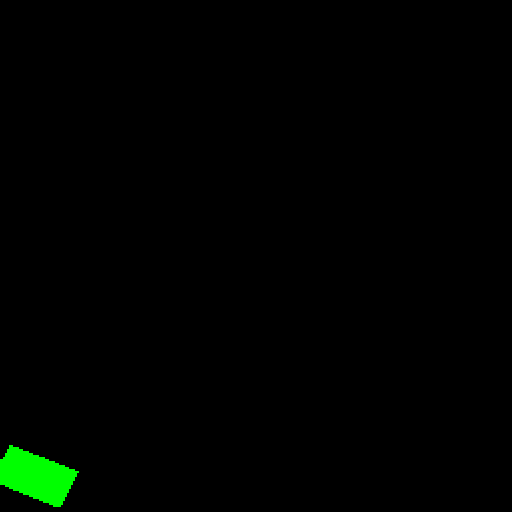}\vspace{2pt}
\end{minipage}}
\subfloat[CDMaskFormer]{
\begin{minipage}[t]{0.12\linewidth}
\includegraphics[width=1\linewidth]{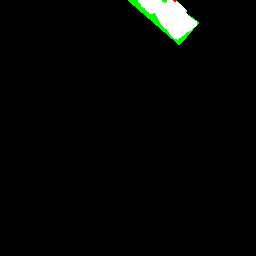}\vspace{2pt}
\includegraphics[width=1\linewidth]{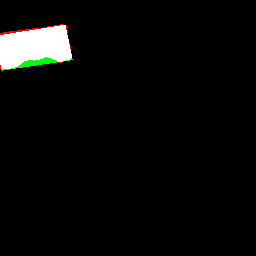}\vspace{2pt}
\includegraphics[width=1\linewidth]{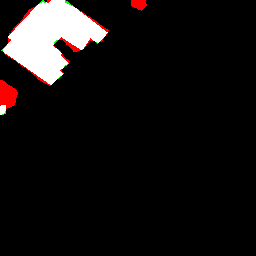}\vspace{2pt}
\includegraphics[width=1\linewidth]{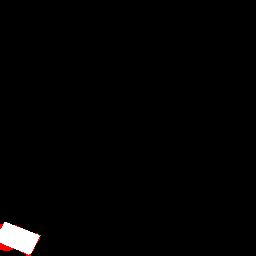}\vspace{2pt}
\end{minipage}}
\caption{Example results output from RSCD methods on test sets from LEVIR-CD dataset. Pixels are colored differently for better visualization (i.e., white for true positive, black for true negative, red for false positive, and green for false negative).}
\label{fig:levir}
\end{figure}

\section{Dataset and Implementation Details.}\label{sec:detail}
\subsection{Dataset}

\textbf{WHU-CD~\cite{whu}} is a building CD dataset. It contains a pair of $32507 \times 15354$-sized dual-time aerial images with a spatial resolution of 0.075 m. Following previous work~\cite{bit}, the paper crops the images into $256\times256$-sized blocks and randomly divides them into a training set (6096 images), a validation set (762 images) and a test set (762 images).

\textbf{LEVIR-CD~\cite{levir}} is a large-scale remotely sensed building change detection dataset. It contains 637 pairs of high-resolution dual-temporal remote sensing images of $1024\times 1024$ size and 0.5 m spatial resolution. Following the default settings of the dataset, the paper crops the images into non-overlapping blocks of $256\times256$ size and randomly divides them into a training set (7120 images), a validation set (1024 images) and a test set (2048 images).

\textbf{DSIFN-CD~\cite{dsifn}} is a high-resolution bi-temporal CD dataset. It contains the change of multiple kinds of land-cover objects, such as roads, buildings, croplands, and water bodies. The paper follows the default cropped samples of size $512\times512$ provided by the authors and the default way to divide them into a training set (3600 images), a validation set (340 images) and a test set (48 images).

\textbf{CLCD~\cite{clcd}} is a cropland change detection dataset containing 600 pairs of remote sensing images of $512\times512$ size with spatial resolution ranging from 0.5 m to 2 m. The paper randomly divides it into a training set (360 images), a validation set (120 images) and a test set (120 images).

\textbf{SYSU-CD~\cite{sysucd}} is a high-resolution bi-temporal change detection dataset that includes 20,000 pairs of orthographic aerial images with a spatial size of $256 \times 256$ and a spatial resolution of 0.5 meters, primarily collected over Hong Kong. The dataset captures a diverse range of land-cover objects, including buildings, vessels, roads, and vegetation, offering a significant challenge for change detection tasks. For data distribution, the dataset is divided into a training set (12,000 images), a validation set (4,000 images), and a test set (4,000 images).

\subsection{Implementation Details}

We implement our CDMaskFormer using Python based on PyTorch library, where a workstation with two NVIDIA GTX A6000 graphics cards and four NVIDIA GTX A5000 graphics cards (192~GB GPU memory in total) is employed. The initial learning rate is $1e-4$ ($5e-4$ for backbone) and the Adam~\cite{adam} optimizer is adopted with a weight decay of $0.05$ ($0.01$ for backbone). During the training period, we employ a poly learning rate decay strategy with the power of $0.9$. The batch size is set to 8 for LEVIR-CD and WHU-CD, while 4 for DSIFN-CD and CLCD. We conduct data augmentation by flipping and bluring the images for training.

\subsection{Evaluation Metrics}

We measure and report Precision (Pre.), Recall (Rec.), Intersection over Union (IoU), and Overall Accuracy (OA) for the change category within the test set. These metrics are defined in the following way:

\begin{equation}
    Pre=\frac{TP}{TP+FP}
\end{equation}
\begin{equation}
    Rec=\frac{TP}{TP+FN}
\end{equation}
\begin{equation}
    IoU=\frac{TP}{TP+FN+FP}
\end{equation}
\begin{equation}
    OA=\frac{TP+TN}{TP+TN+FN+FP}
\end{equation}
where TP, TN, FP, and FN represent the number of true positive, true negative, false positive, and false negative, respectively.

We use the F1-score (F1) of the change category as the primary evaluation metric within the test set. The F1-score is derived from the precision and recall of the test set, calculated using the harmonic mean of these two metrics as follows:

\begin{equation}
    F1=\frac{2 \times (Rec \times Pre)}{Rec + Pre}
\end{equation}

\begin{figure}[t]
\centering
\captionsetup[subfloat]{labelsep=none,format=plain,labelformat=empty,font=tiny}
\subfloat[$T_1$]{
\begin{minipage}[t]{0.12\linewidth}
\includegraphics[width=1\linewidth]{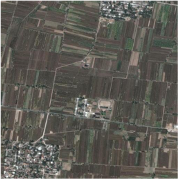}\vspace{2pt}
\includegraphics[width=1\linewidth]{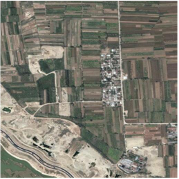}\vspace{2pt}
\includegraphics[width=1\linewidth]{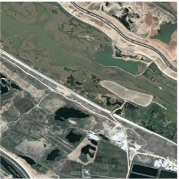}\vspace{2pt}
\includegraphics[width=1\linewidth]{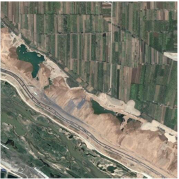}\vspace{2pt}
\end{minipage}}
\subfloat[$T_2$]{
\begin{minipage}[t]{0.12\linewidth}
\includegraphics[width=1\linewidth]{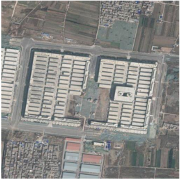}\vspace{2pt}
\includegraphics[width=1\linewidth]{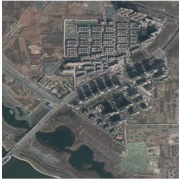}\vspace{2pt}
\includegraphics[width=1\linewidth]{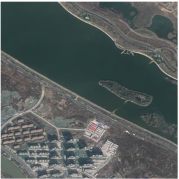}\vspace{2pt}
\includegraphics[width=1\linewidth]{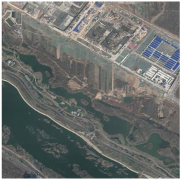}\vspace{2pt}
\end{minipage}}
\subfloat[GT]{
\begin{minipage}[t]{0.12\linewidth}
\includegraphics[width=1\linewidth]{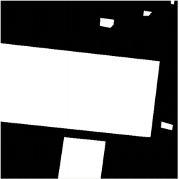}\vspace{2pt}
\includegraphics[width=1\linewidth]{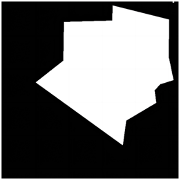}\vspace{2pt}
\includegraphics[width=1\linewidth]{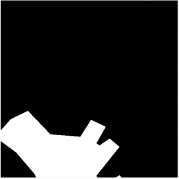}\vspace{2pt}
\includegraphics[width=1\linewidth]{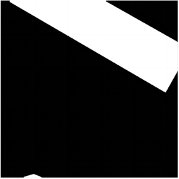}\vspace{2pt}
\end{minipage}}
\subfloat[SNUNet]{
\begin{minipage}[t]{0.12\linewidth}
\includegraphics[width=1\linewidth]{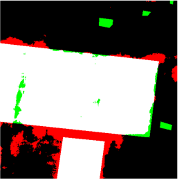}\vspace{2pt}
\includegraphics[width=1\linewidth]{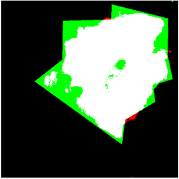}\vspace{2pt}
\includegraphics[width=1\linewidth]{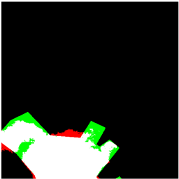}\vspace{2pt}
\includegraphics[width=1\linewidth]{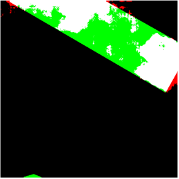}\vspace{2pt}
\end{minipage}}
\subfloat[BIT]{
\begin{minipage}[t]{0.12\linewidth}
\includegraphics[width=1\linewidth]{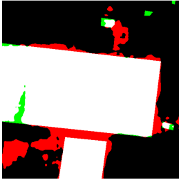}\vspace{2pt}
\includegraphics[width=1\linewidth]{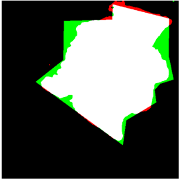}\vspace{2pt}
\includegraphics[width=1\linewidth]{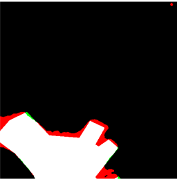}\vspace{2pt}
\includegraphics[width=1\linewidth]{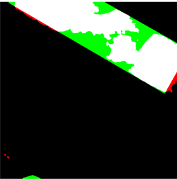}\vspace{2pt}
\end{minipage}}
\subfloat[SARASNet]{
\begin{minipage}[t]{0.12\linewidth}
\includegraphics[width=1\linewidth]{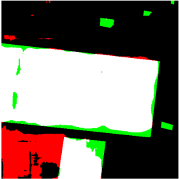}\vspace{2pt}
\includegraphics[width=1\linewidth]{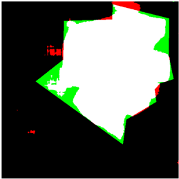}\vspace{2pt}
\includegraphics[width=1\linewidth]{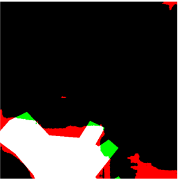}\vspace{2pt}
\includegraphics[width=1\linewidth]{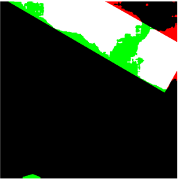}\vspace{2pt}
\end{minipage}}
\subfloat[CDMaskFormer]{
\begin{minipage}[t]{0.12\linewidth}
\includegraphics[width=1\linewidth]{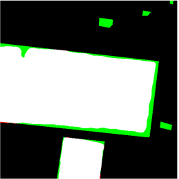}\vspace{2pt}
\includegraphics[width=1\linewidth]{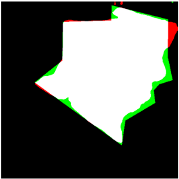}\vspace{2pt}
\includegraphics[width=1\linewidth]{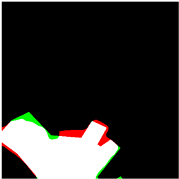}\vspace{2pt}
\includegraphics[width=1\linewidth]{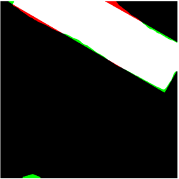}\vspace{2pt}
\end{minipage}}
\caption{Example results output from RSCD methods on test sets from DSIFN-CD dataset. Pixels are colored differently for better visualization (i.e., white for true positive, black for true negative, red for false positive, and green for false negative).}
\label{fig:res-dsifn}
\end{figure}

\begin{figure*}[t]
	\centering
        \includegraphics[width=0.9\textwidth]{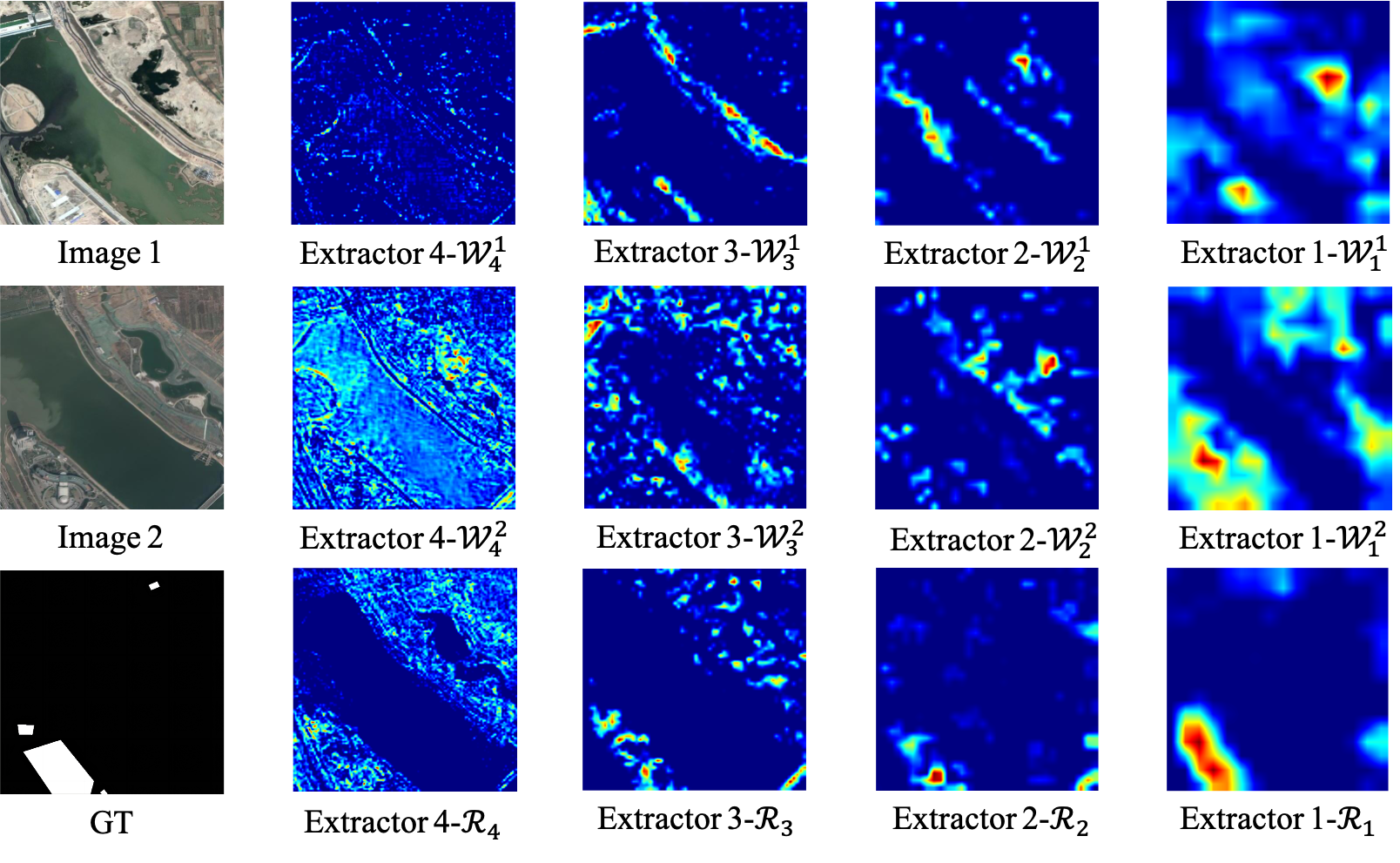}
	\centering
	\caption{Class activation maps for features of change extractors in different layers. Example images are from the DSIFN-CD test set. Extractor 4, Extractor 3, Extractor 2, and Extractor 1 correspond to 1/4, 1/8, 1/16, and 1/32 resolution of the input image, respectively.}
\label{fig:cam_dsifn}
\end{figure*}

\begin{figure*}[t]
	\centering
        \includegraphics[width=0.9\textwidth]{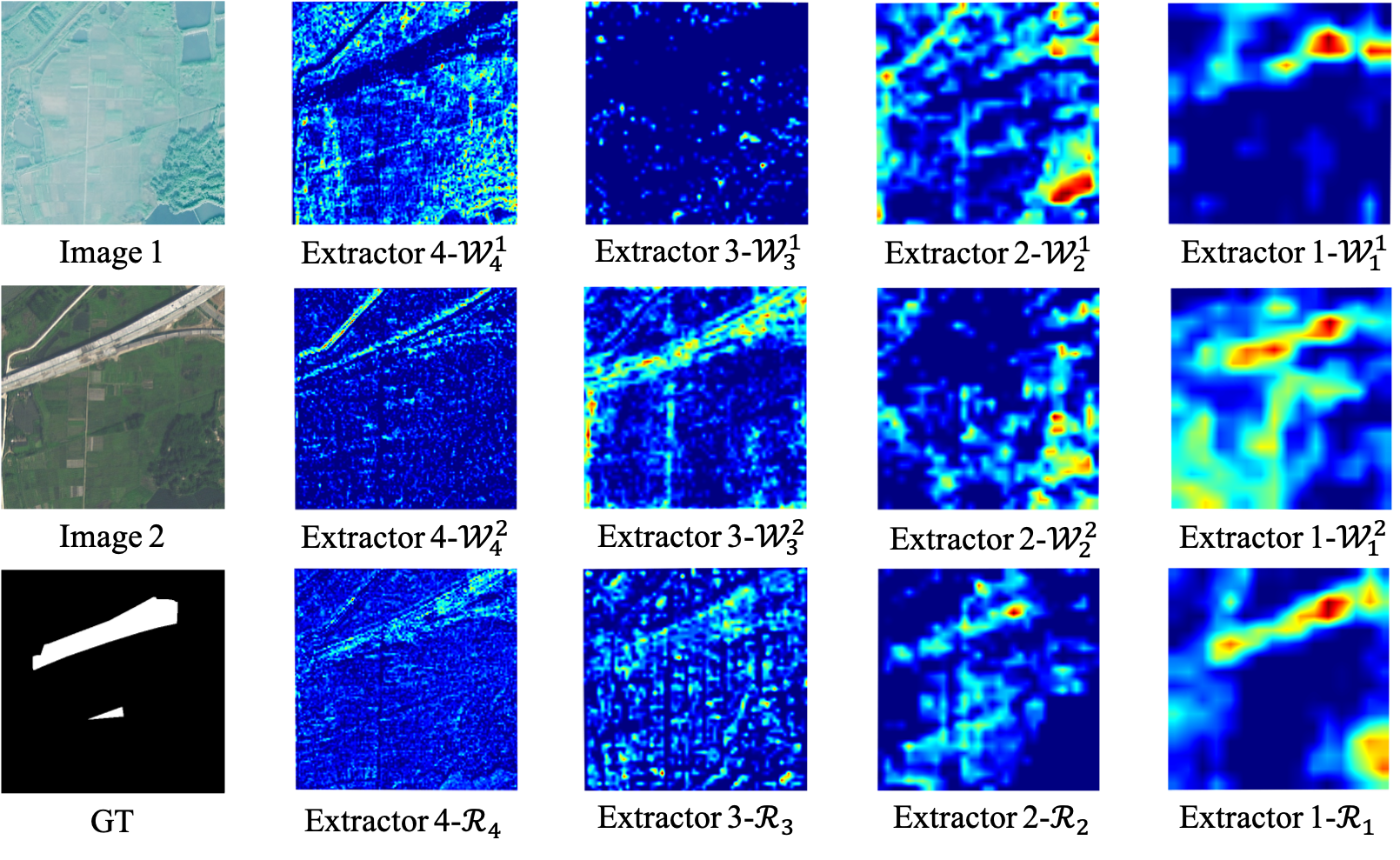}
	\centering
	\caption{Class activation maps for features of change extractors in different layers. Example images are from the CLCD test set. Extractor 4, Extractor 3, Extractor 2, and Extractor 1 correspond to 1/4, 1/8, 1/16, and 1/32 resolution of the input image, respectively.}
\label{fig:cam_clcd}
\end{figure*}

\section{More qualitative visualization.}\label{sec:vis}

We first visualize and compare with some SoTA methods, as shown in Fig~\ref{fig:res-dsifn}. CDMaskFormer significantly reduces FP and FN in complex scenarios. In particular, compared with other methods, CDMaskFormer can detect the boundary of the changing region more clearly. Qualitative analysis proves the effectiveness of CDMaskFormer.
We also perform visual comparisons on the WHU-CD, CLCD and LEVIR-CD datasets. As shown in Fig. \ref{fig:whu}, Fig. \ref{fig:clcd} and Fig. \ref{fig:levir}, the proposed CDMaskFormer demonstrate better visualization performance compared to previous state-of-the-art methods such as SNUNet, BiT, and SARASNet. Specifically, our CDMaskFormer appears to have fewer false alarms (e.g., the fourth row of Fig. \ref{fig:whu} and the first row of Fig. \ref{fig:clcd}) and omission alarms (e.g., the third row of Fig. \ref{fig:whu} and the second row of Fig. \ref{fig:clcd}, the first row of Fig. \ref{fig:levir}). In addition, CDMaskFormer outputs masks with sharper boundaries (e.g., the third row of Fig. \ref{fig:clcd}) as well as more complete topological shapes (e.g., the second row of Fig. \ref{fig:whu} and the fourth row of Fig. \ref{fig:levir}). These further validate our CDMaskFormer's capability in effectively enhancing interest changes and capture change region details in a variety of complex scenarios.

To explore the validity of the model, we additionally visualize the activation maps of the features output by the change extractor on the DSIFN-CD and CLCD datasets, which is implemented based on Grad-CAM. As shown in Fig. \ref{fig:cam_dsifn} and Fig. \ref{fig:cam_clcd}, the activation values of the change extractor in the change region gradually increase as the depth of the model deepens, which proves that the model is effective in enhancing the semantic feature differences in the change region.

\section{CDPixel analysis}\label{sec:cdpixel}
We provide a detailed description for Fig.~\ref{fig:intro}, where BIT (a classic instance network of CDPixel) is experimented on the DSIFN dataset. Specifically, after we have trained BIT on the training set, we select several images on the test set for visualization of the features in the changed area. It can be observed that the features in the changed area on different images are far away from each other, even on the same image. This can be explained by the fact that the complexity of the scene and variations in imaging conditions (e.g., weather, light, season, and frequent irrelevant changes caused by human actions) in the scene, both of which lead to a large variance in the features of the changing pixels. As a result, previous studies (i.e., CDPixel) are often built on fixed change prototypes obtained from the training set and used for the test set, resulting in poor performance due to intolerance of such rich and diverse changes. This finding motivates us to propose CDMask, which introduces learnable change queries that adaptively update and predict a set of binary masks based on bi-temporal image feature content, and classify the masks to determine whether a change of interest has occurred. As a result, CDMask is able to adapt to different latent data distributions and thus accurately identify interest change regions in complex scenes.

\section{Limitation analysis}\label{sec:limit}
Although the effectiveness of CDMaskFormer has been verified by extensive experiments, there are still two aspects that could be improved: 1) a more efficient component design. CDMaskFormer introduces a highly customizable change extractor and transformer decoder, but the pixel decoder is still instantiated using basic deformable attention \cite{deformabledetr}. In addition, it remains to be explored whether various optimization strategies for DETRs, such as ground-truth mask pilote \cite{mpformer}, one-to-many matching \cite{groupdetr}, and contrastive denoising Training \cite{dino}, can further improve the performance of CDMask's instantiation network.
2) Establishment of Unified Architecture. CDMask has verified that change detection tasks can be handled using DETR-like models with minor architectural changes. Therefore, it also provides a promising research target that build a truly unified architecture for remote sensing image segmentation and change detection.

\end{document}